%% file: main.tex
\documentclass[10pt,twocolumn,letterpaper]{article}

\usepackage{iccv}              

\input{preamble}
\definecolor{iccvblue}{rgb}{0.21,0.49,0.74}
\usepackage[pagebackref,breaklinks,colorlinks,allcolors=iccvblue]{hyperref}

\def\paperID{*****} 
\def\confName{ICCV}
\def\confYear{2025}
\usepackage{multirow}
\usepackage[pagebackref,breaklinks,colorlinks,allcolors=iccvblue]{hyperref}
\usepackage{pifont}
\usepackage{makecell}
\usepackage{subcaption} 
\usepackage{diagbox}

\title{Human-Inspired Summarization: Cluster Scene Videos into Diverse Frames}

\author{Chao Chen$^{1}$,\, Mingzhi Zhu$^{1,}\thanks{Equal contribution}$,\, Ankush Pratap Singh$^{1,}\footnotemark[1]$,\, Yu Yan$^{1}$,\, Felix Juefei-Xu$^{1}$, Chen Feng$^{1,}$\thanks{The corresponding author is Chen Feng {\tt\small cfeng@nyu.edu}}\\
$^{1}$New York University \\
}

\begin{document}
\maketitle
\input{sections/0-title-author-abstract}
\input{sections/1-introduction}
\input{sections/2-related}

\input{sections/3-method}
\input{sections/4-experiments}
\input{sections/5-conclusion}
{
    \small
    \bibliographystyle{ieeenat_fullname}
    \bibliography{main}
}

\input{sections/6-supplementary}
\end{document}

%% file: sections/0-title-author-abstract.tex
\begin{abstract}
Humans are remarkably efficient at forming spatial understanding from just a few visual observations. When browsing real estate or navigating unfamiliar spaces, they intuitively select a small set of views that summarize the spatial layout. Inspired by this ability, we introduce scene summarization, the task of condensing long, continuous scene videos into a compact set of spatially diverse keyframes that facilitate global spatial reasoning. Unlike conventional video summarization—which focuses on user-edited, fragmented clips and often ignores spatial continuity—our goal is to mimic how humans abstract spatial layout from sparse views. We propose SceneSum, a two-stage self-supervised pipeline that first clusters video frames using visual place recognition to promote spatial diversity, then selects representative keyframes from each cluster under resource constraints. When camera trajectories are available, a lightweight supervised loss further refines clustering and selection. Experiments on real and simulated indoor datasets show that SceneSum produces more spatially informative summaries and outperforms existing video summarization baselines.

\end{abstract}

%% file: sections/1-introduction.tex
\vspace{-4mm}
\section{Introduction}
\label{sec:intro}

A scene summary, represented by a set of images, supports applications like surveillance and real estate platforms (e.g., Zillow, Google Maps), providing a concise visual overview that helps users quickly grasp space, layout, and ambiance without detailed inspection or visits. Humans achieve this effortlessly by selectively integrating sparse, key views to build a coherent spatial understanding. Inspired by this human ability, an effective summary should cover as many distinct areas as possible within a limited image budget.

\begin{figure}[t]
\vspace{-10mm}
    \centering {\includegraphics[width=0.47\textwidth
    ]{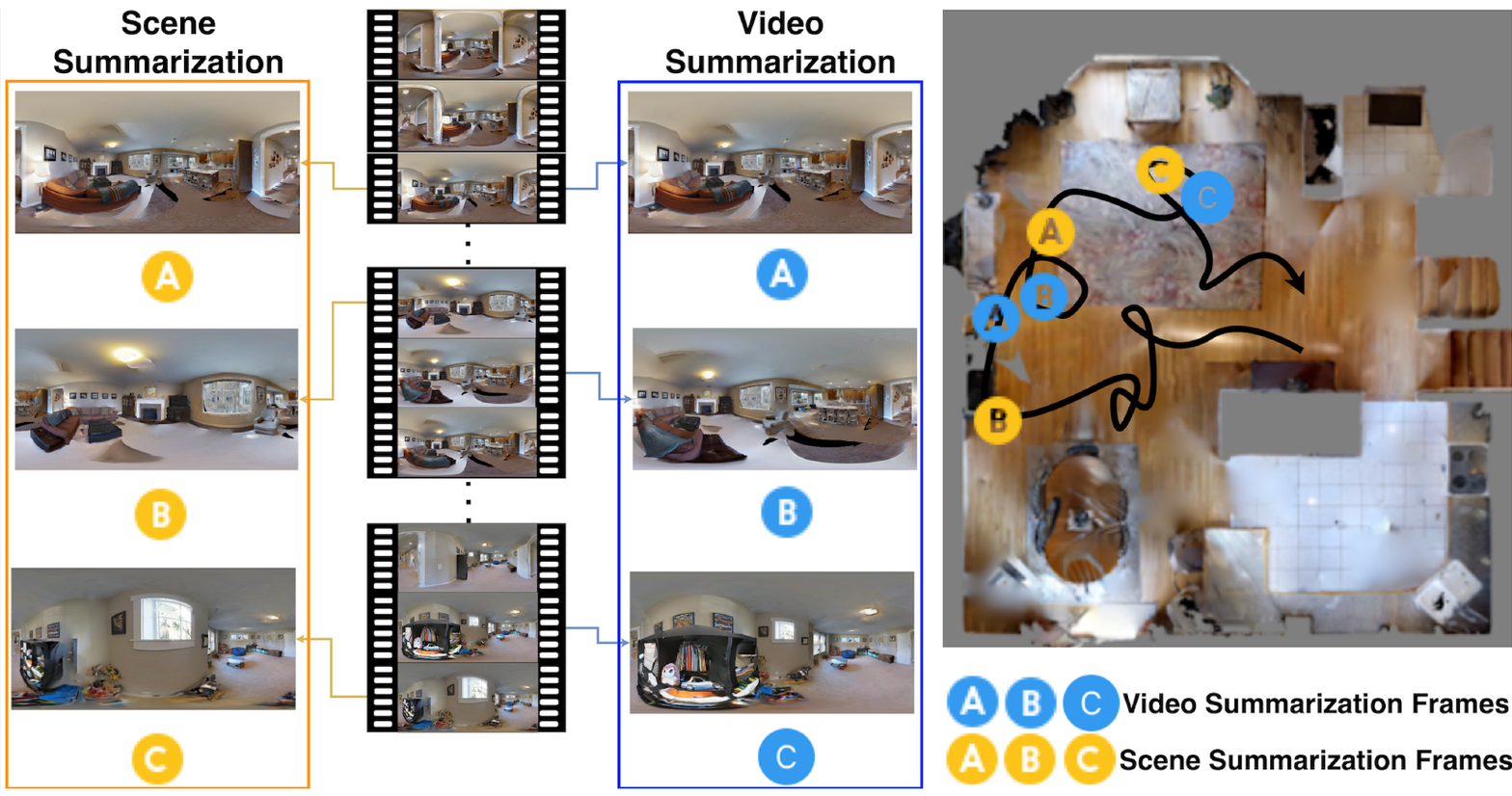}} \quad
    \vspace{-16mm}
    \caption{Scene summarization vs video summarization: orange box shows frames selected by scene summarization, blue box shows those by video summarization. The right side displays the camera trajectory and floorplan, with blue/orange nodes indicating video/scene summarization frames. We observe that scene summarization promotes \textit{spatial} diversity in the summarized frames, unlike video summarization that may select frames in close spatial proximity (\textcolor[HTML]{007FFF}{frame A and B}).} Scene summarization aligns with how humans intuitively select spatially diverse key views for efficient scene understanding.
    \label{fig:teasing}
\vspace{-6mm}
\end{figure}

 To obtain such a summary, classic video summarization aims to compress video clips based on factors like color or temporal distribution. However, it was originally designed for artistically edited, discontinuous video clips, and overlooks scene coverage due to their fragmented nature. Unlike humans who select spatially diverse views to build holistic understanding, it neglects both \textit{spatial diversity} and the \textit{holistic understanding} of the environment in its summarized frames. As shown in Fig.~\ref{fig:teasing}, these frames are frequently spatially proximate to each other, posing a significant challenge in the scene summarization task.

We introduce scene summarization as a novel, human-inspired task at the intersection of robotics, computer vision, and multimedia analysis. It differs from video summarization in several key ways, primarily in terms of \textit{temporal}, \textit{color}, and \textit{spatial} characteristics. The core goal is to condense lengthy scene videos into a concise set of keyframe images, each associated with a geographical coordinate to capture spatial information. This process gains particular importance when autonomous robots continually explore and capture \textit{redundant} images of the same environments. In such scenarios, scene summarization enables robots to distinguish, condense, concisely describe, and comprehend the important visual aspects of the environment they encounter. 

An intuitive approach to addressing the scene summarization task involves adopting video summarization techniques. There have been conventional techniques, such as a color-based scene change detection python library PySceneDetect~\cite{pyscenedetect2017}, and contemporary deep learning approaches like DR-DSN~\cite{zhou2018deep} and CA-SUM~\cite{apostolidis2022summarizing} which aims to condense extensive video content while preserving visual keyframes. However, these conventional techniques fail to utilize \textit{spatial information} and do not provide a \textit{holistic understanding} of the video as a continuous trajectory. 


To enhance spatial sparsity and holistic understanding in keyframes, we propose two strategies:

\textbf{Spatially-Informed feature clustering.} Clustering based methods like VSUMM~\cite{de2011vsumm} group datasets without considering spatial context. Inspired by how humans integrate spatial structure when summarizing scenes, we incorporate spatial information into feature clustering. This allows the model to capture spatial relationships and group features from coherent regions, such as grouping features from the same room together.

\textbf{Selecting keyframes without feature centroids.} Traditional methods select keyframes based on feature centroids~\cite{de2011vsumm}, which often fail to correspond to actual spatial centers, reducing accuracy. Inspired by human ability to distinguish and focus on salient and spatially meaningful views rather than simple averages, we propose using contrastive learning to increase the separation between feature centroids of different clusters, enhancing cluster distinction. However, processing all images within large clusters raises \textit{GPU memory challenges}. To overcome this, we design a sampling-based model that efficiently learns the key representative features of each cluster, mimicking how humans selectively focus on key observations in complex scenes.

We propose \textbf{SceneSum}, a two-stage, human-inspired framework for scene summarization. The first stage, \textit{clustering}, mimics how humans segment spatial regions by grouping long trajectories using spatially-informed VPR features, outperforming classic contrastive clustering. The second stage, \textit{keyframe selection}, models human attention by selecting one representative keyframe per cluster via a sampling-based masked autoencoder with contrastive loss, efficiently retrieving $K$ spatially central keyframes without memory issues. Our main contributions are:

\begin{enumerate}
    \item We propose a scene summarization task focused on spatially diverse keyframes to produce more concise and varied video summaries.
    \item We introduce \textit{SceneSum}, a two-stage framework where the first stage segments a trajectory into clusters and the second selects one frame per cluster using a masked autoencoder-decoder variant. Additionally, We demonstrate that our self-supervised framework can easily be adapted to a supervised method by integrating ground truth for keyframe selection. Our approach outperforms common video summarization methods in scene summarization.
    \item We perform an ablation study showing VPR-based features cluster spatially close frames better than classic contrastive features, leading to improved scene summarization.

\end{enumerate}

%% file: sections/2-related.tex
\section{Related Work}
\begin{figure*}[t]
\vspace{-2mm}
    \centering {\includegraphics[width=0.88\textwidth
    ]{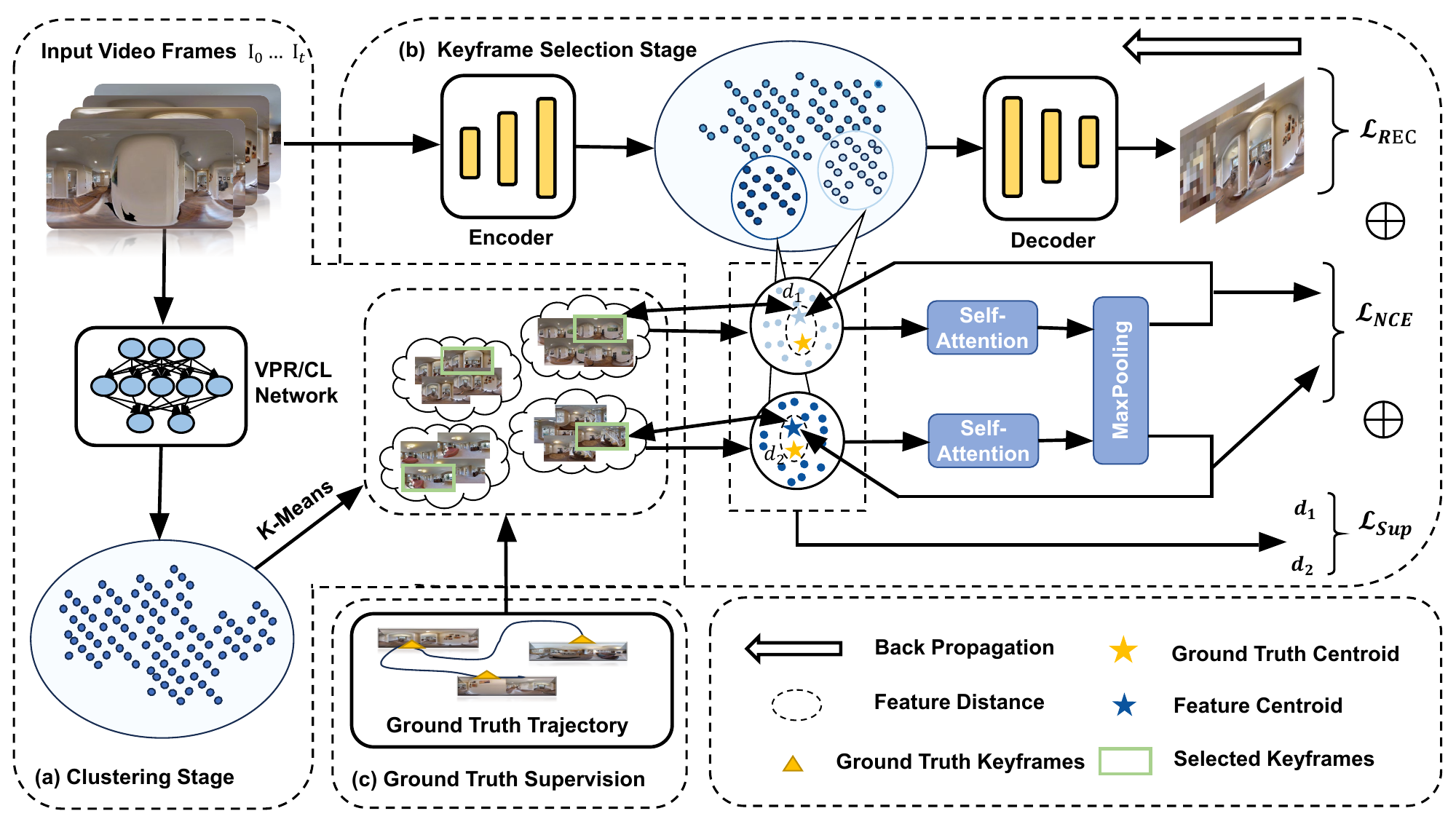}} \quad
\vspace{-4mm}
    \caption{\textbf{Overview of SceneSum.} Our approach consists of two main stages: (1) Clustering: we use a contrastive learning (CL) or visual place recognition (VPR) encoder to encode images into features, then cluster frames based on these features or ground truth odometry if available. (2) Keyframe Selection: we select the most representative and visually distinct frame from each cluster. If spatial information is available, an optional Ground Truth supervision stage can be enabled, switching the model from self-supervised to supervised by using pre-computed keyframes from ground truth trajectories to guide keyframe selection. Frames circled in \textcolor{teal}{green} denote the selected keyframes.}
    \label{fig:workflow}
\vspace{-3mm}

\end{figure*}

\textbf{Classic video summarization}. A range of classic techniques has emerged, falling into categories such as clustering-based, change detection-based, and sparse-dictionary-based methods. Within clustering-based approaches, some studies have specifically focused on utilizing color histograms~\cite{de2011vsumm,ejaz2012adaptive,furini2010stimo,wu2017novel,mundur2006keyframe}. For instance, VSUMM~\cite{de2011vsumm} utilizes K-means clustering, designating each cluster centroid as a keyframe. Meanwhile, STIMO~\cite{furini2010stimo} employs the Farthest Point-First (FPF) technique to analyze the color distribution of each frame, dynamically creating video storyboards. 
Change detection-based methods measure the complexity of the sequence in terms of changes in the visual content~\cite{gianluigi2006AnIA,ejaz2012adaptive,liang2021news,doulamis2000fuzzy,srinivas2016improved}. Gianluigi and Raimondo~\cite{gianluigi2006AnIA} introduce a method for choosing representative key frames utilizing various frame descriptors. In contrast, VSUKFE~\cite{ejaz2012adaptive} relies on inter-frame discrepancies computed from RGB color channel correlation, color histogram, and moments of inertia. Lastly, one of the proposed techniques uses sparse modeling for representative selection through convex optimization~\cite{elhamifar2012see}.

\textbf{Supervised video summarization}. 
Deep learning-based video summarization leverages advanced neural network architectures (CNN~\cite{rochan2018video,chu2019spatiotemporal,elfeki2019video}, attention~\cite{fajtl2019summarizing,ji2019video,ji2020deep,li2021exploring,liu2019learning}, GAN~\cite{mahasseni2017unsupervised,yuan2019cycle,apostolidis2020ac,jung2019discriminative,jung2020global}, and LSTM~\cite{zhao2018hsa,zhang2016video}) to automatically extract salient information from videos. Among those, supervised approaches typically rely on human annotations of keyframes within a video~\cite{rochan2018video,chu2019spatiotemporal,elfeki2019video,fajtl2019summarizing}. HSA-RNN~\cite{zhao2018hsa} is one typical hierarchical structure-adaptive RNN model that focuses on the temporal dependency of video frames~\cite{zhang2016video,zhao2018hsa,ji2019video,ji2020deep}. In contrast, there are several works~\cite{chu2019spatiotemporal,elfeki2019video,huang2019novel} focusing on the spatial-temporal structure of video frames. Notably, iPTNet~\cite{jiang2022joint} introduces an important propagation-based network for joint video summarization, enhancing training through cross-task sample transfer from related datasets.

\textbf{Self-supervised video summarization}. 
Within the realm of self-supervised techniques~\cite{yang2015unsupervised,mahasseni2017unsupervised,jung2020global,apostolidis2022summarizing,zhang2018unsupervised}, several prominent methods have emerged, including adversarial learning-based~\cite{mahasseni2017unsupervised,yuan2019cycle,apostolidis2020ac,jung2019discriminative,jung2020global,rochan2019video}, reinforcement learning-based~\cite{zhou2018deep,zhao2019property,yaliniz2021using}, motion-based~\cite{zhang2020unsupervised} approaches.  In the domain of adversarial learning, SUM-GAN~\cite{mahasseni2017unsupervised,yuan2019cycle} is widely used in unsupervised video summarization task to fool the discriminator when seeing the machine and human-generated summaries. Comparably, reinforcement learning approaches aim to achieve predefined summary properties. For instance, DSN~\cite{zhou2018deep} assigns probabilities to video frames, indicating their selection likelihood. It then utilizes these probability distributions to make frame selections, ultimately generating video summaries. Lastly, motion-based approaches~\cite{zhang2020unsupervised} use stacked sparse LSTM auto-encoders to capture key motions of important visual objects.

\textbf{SLAM-based summarization}. Video summarization using SLAM estimates frame poses and selects keyframes based on these estimates. In outdoor environments, methods like Zhang~\cite{zhang2012multi} use GPS to predict trajectories and identify hotspots, though GPS coverage can miss some popular spots due to the urban canyon effect~\cite{stalbaum2013keyframe}. In indoor scenarios, the lack of GPS makes prediction difficult, leading to errors and pose drifting~\cite{yang2019amobile,das2015entropy}. Besides, Yang~\cite{yang2019amobile} suggests following a person for video capture, but this may not work well in empty rooms. While SLAM-based summarization is straightforward, it is prone to losing track.

\textbf{Clustering}. The goal of clustering is to partition a set of data points into distinct groups or clusters. Most video summarization methods~\cite{zhou2019automatic,abualigah2018new,abualigah2018hybrid} utilize clustering to group similar frames together, reducing data complexity, eliminating redundancy, and extracting key scenes for efficient summary generation. In this paper, we select several clustering methods along with K-Nearest Neighbors (KNN)~\cite{peterson2009k} as baselines: contrastive-based clustering methods~\cite{li2020prototypical,chen2020simple,he2020momentum}, and VPR-based clustering methods~\cite{arandjelovic2016netvlad,hausler2021patch,ali2023mixvpr}, because of their applications in visual navigation in~\cite{kwon2021visual,chen2023deepmapping2}. Importantly, these clustering methods reflect human-like scene discrimination by grouping spatially coherent views, thus better mimicking how humans segment and understand environments.

%% file: sections/3-method.tex
\providecommand{\mo}{\mathbf{o}}

\section{Scene Summarization}

\subsection{Problem Formulation}\label{sec:overview}

\begin{figure*}[t]
     \centering
     \setlength{\abovecaptionskip}{4pt}
     \begin{subfigure}[t]{0.2\textwidth}
         \centering
         \includegraphics[width=\textwidth]{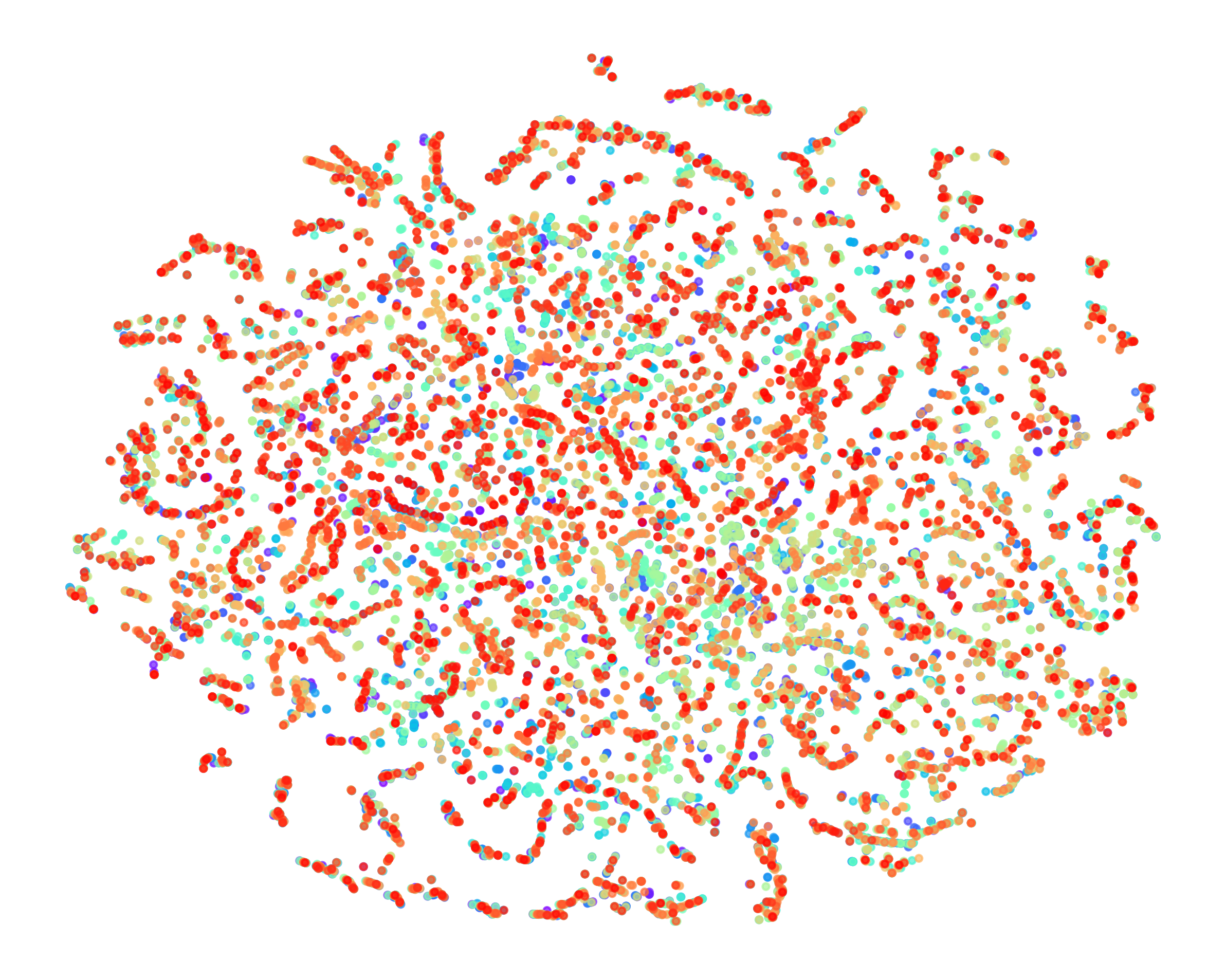}
         \vspace{-5mm}
         \caption{zero-shot MoCo}
         \vspace{-2mm}
         \label{fig:MoCo_cluster}
     \end{subfigure}
     \begin{subfigure}[t]{0.2\textwidth}
         \centering
         \includegraphics[width=\textwidth]{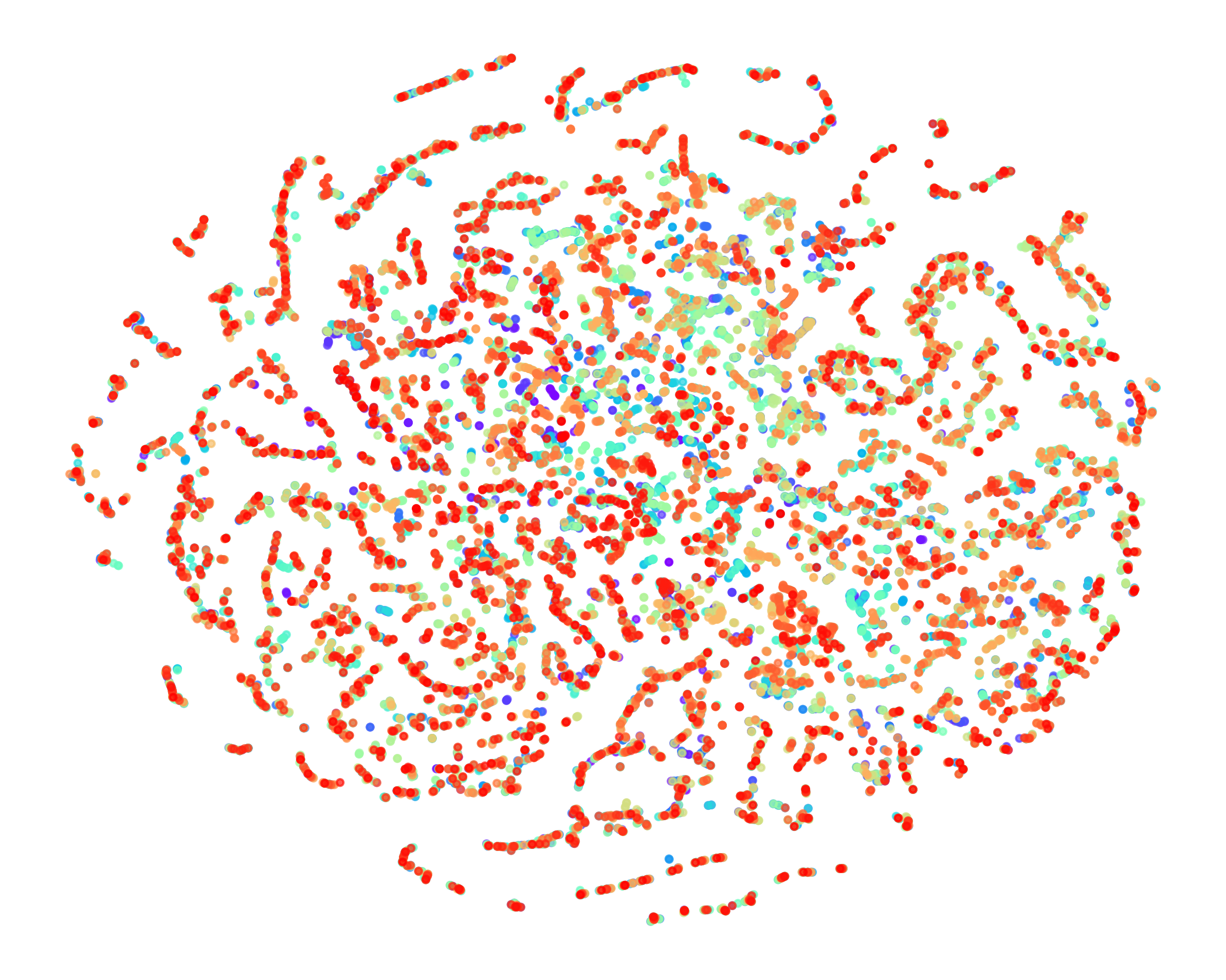}
         \vspace{-5mm}
         \caption{zero-shot SimCLR}
         
         \label{fig:simclr_cluster}
     \end{subfigure}
     \begin{subfigure}[t]{0.2\textwidth}
         \centering
         \includegraphics[width=\textwidth]{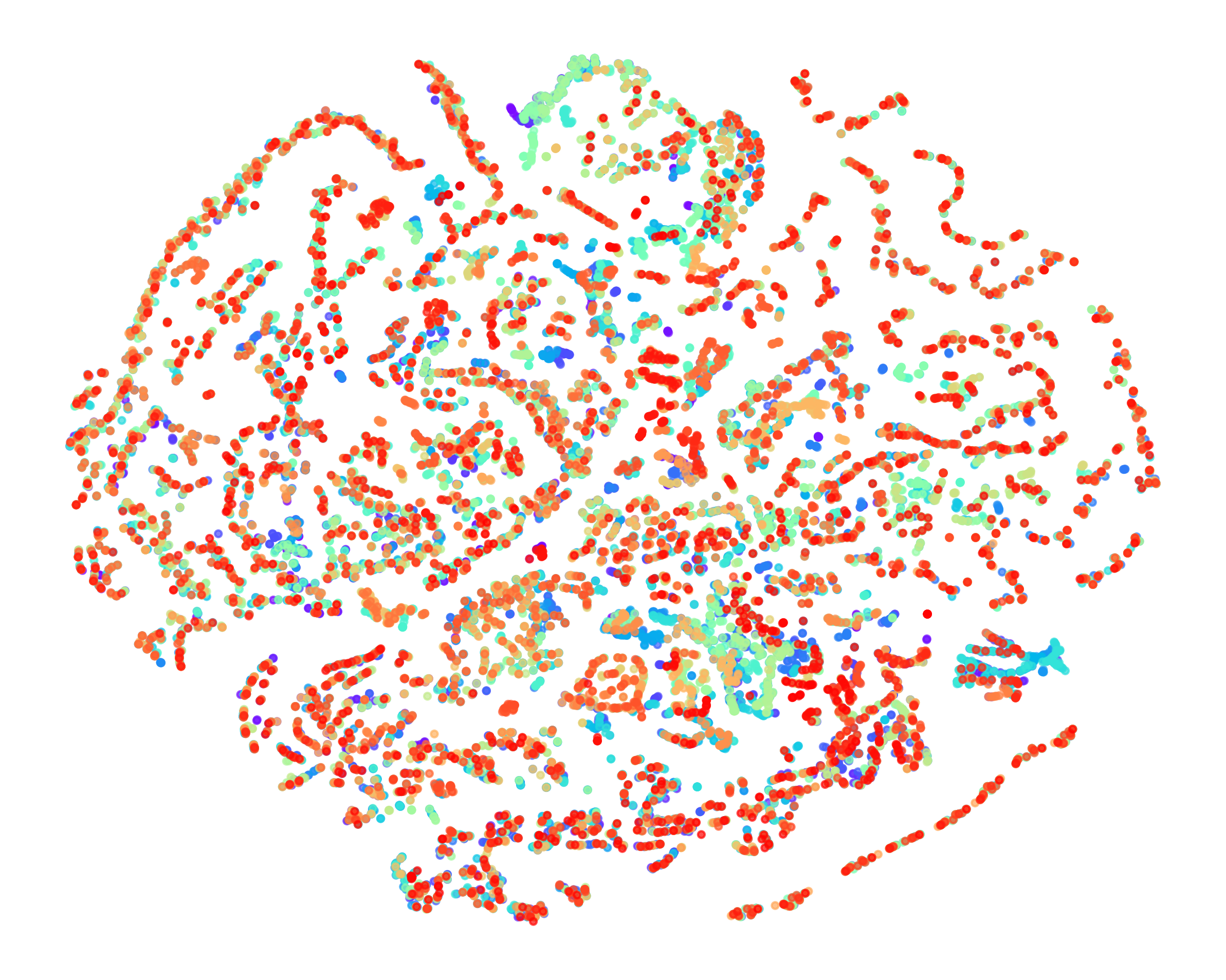}
         \vspace{-5mm}
         \caption{zero-shot PCL}
         
         \label{fig:pcl_cluster}
     \end{subfigure}
     \begin{subfigure}[t]{0.2\textwidth}
         \centering
         \includegraphics[width=\textwidth]{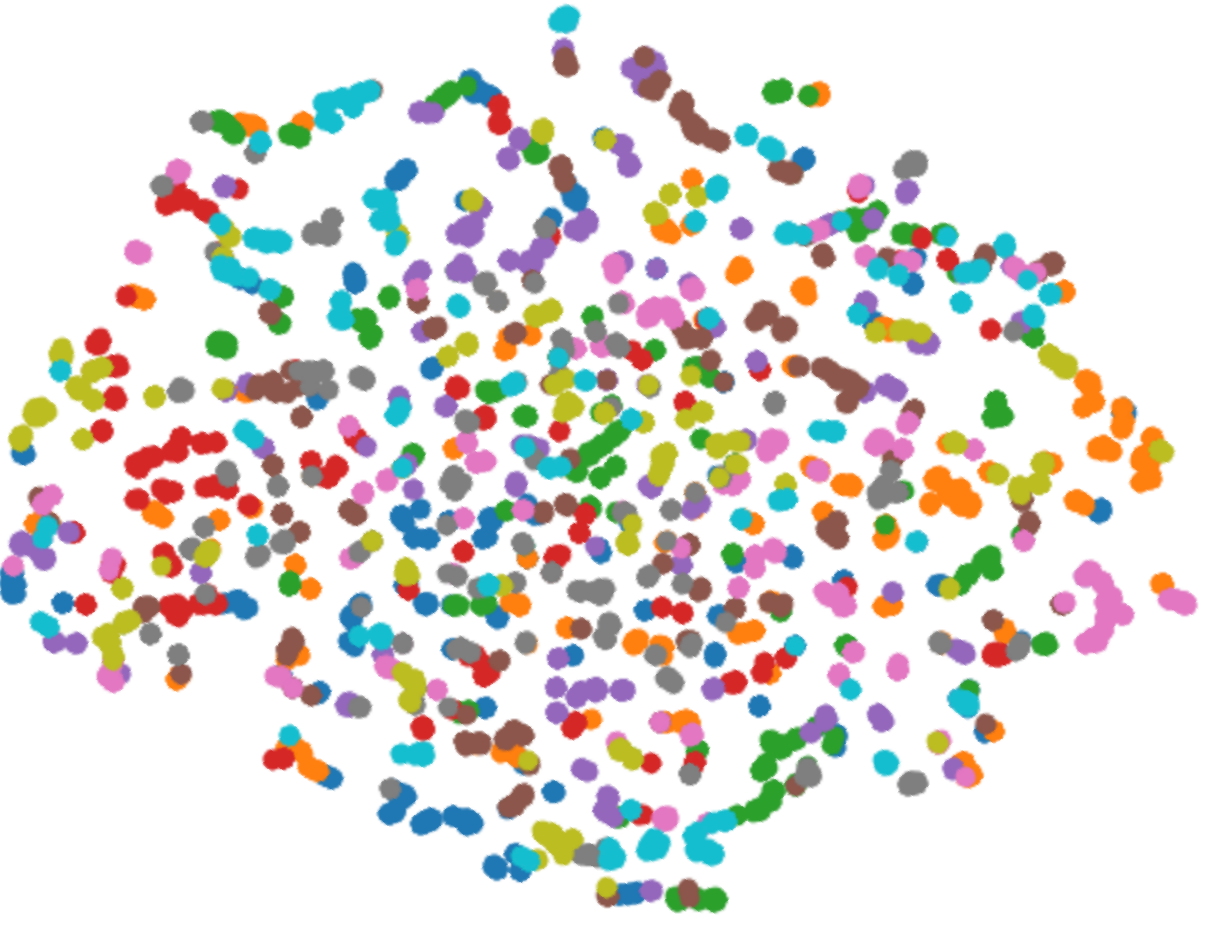}
         \vspace{-5mm}
         \caption{Fine-tuned PCL}
         
         \label{fig:ft_pcl_cluster}
     \end{subfigure}
     \begin{subfigure}[t]{0.2\textwidth}
         \centering
         \includegraphics[width=\textwidth]{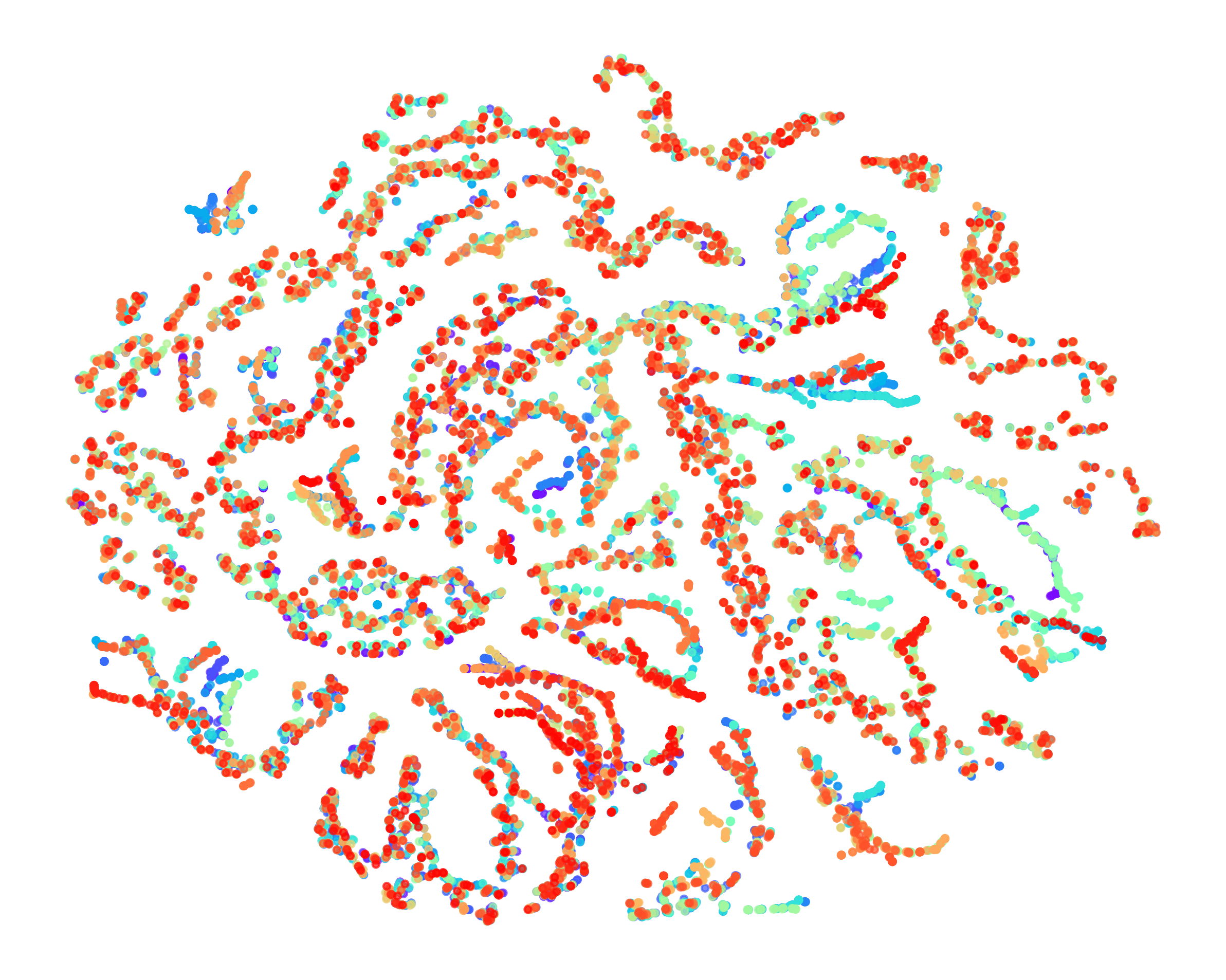}
          \vspace{-5mm}
         \caption{zero-shot Patch-NetVLAD}
         
         \label{fig:patch_cluster}
     \end{subfigure}
     \begin{subfigure}[t]{0.2\textwidth}
         \centering
         \includegraphics[width=\textwidth]{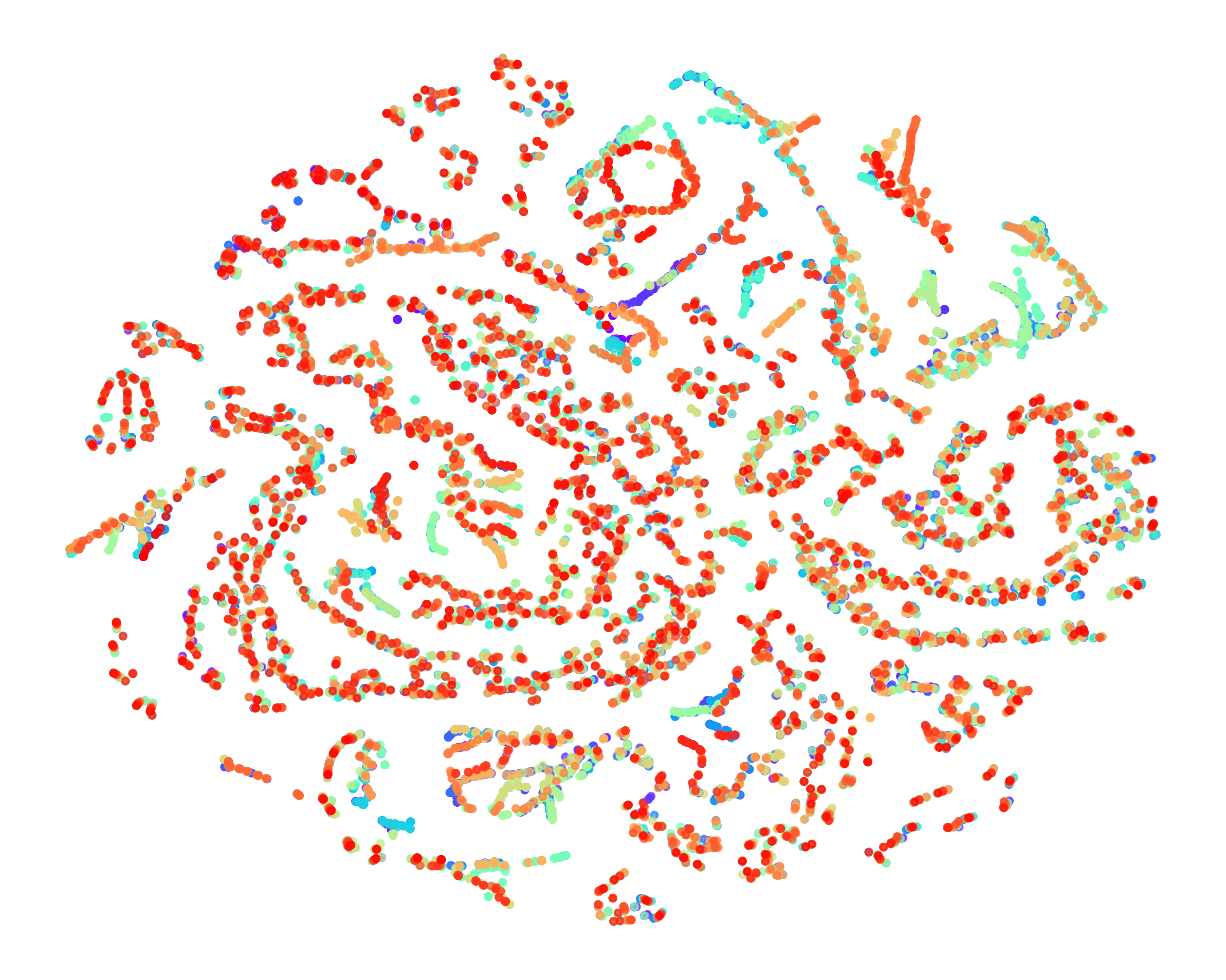}
         \vspace{-5mm}
         \caption{zero-shot NetVLAD}
         
         \label{fig:netvlad_cluster}
     \end{subfigure}
     \begin{subfigure}[t]{0.2\textwidth}
         \centering
         \includegraphics[width=\textwidth]{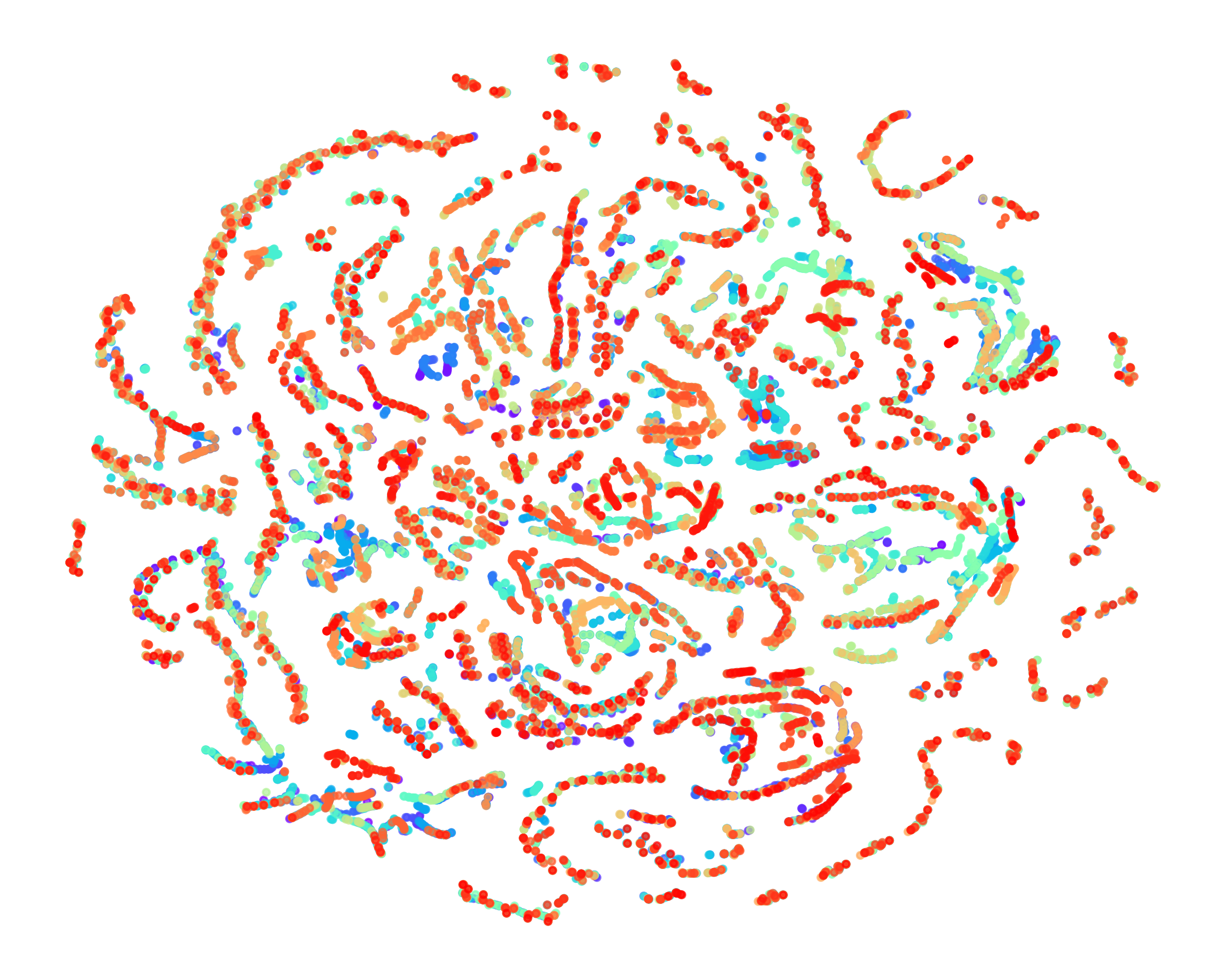}
         \vspace{-5mm}
         \caption{zero-shot MixVPR}
         
         \label{fig:mixvpr_cluster}
     \end{subfigure}
     \begin{subfigure}[t]{0.2\textwidth}
        \centering
        \includegraphics[width=\textwidth]{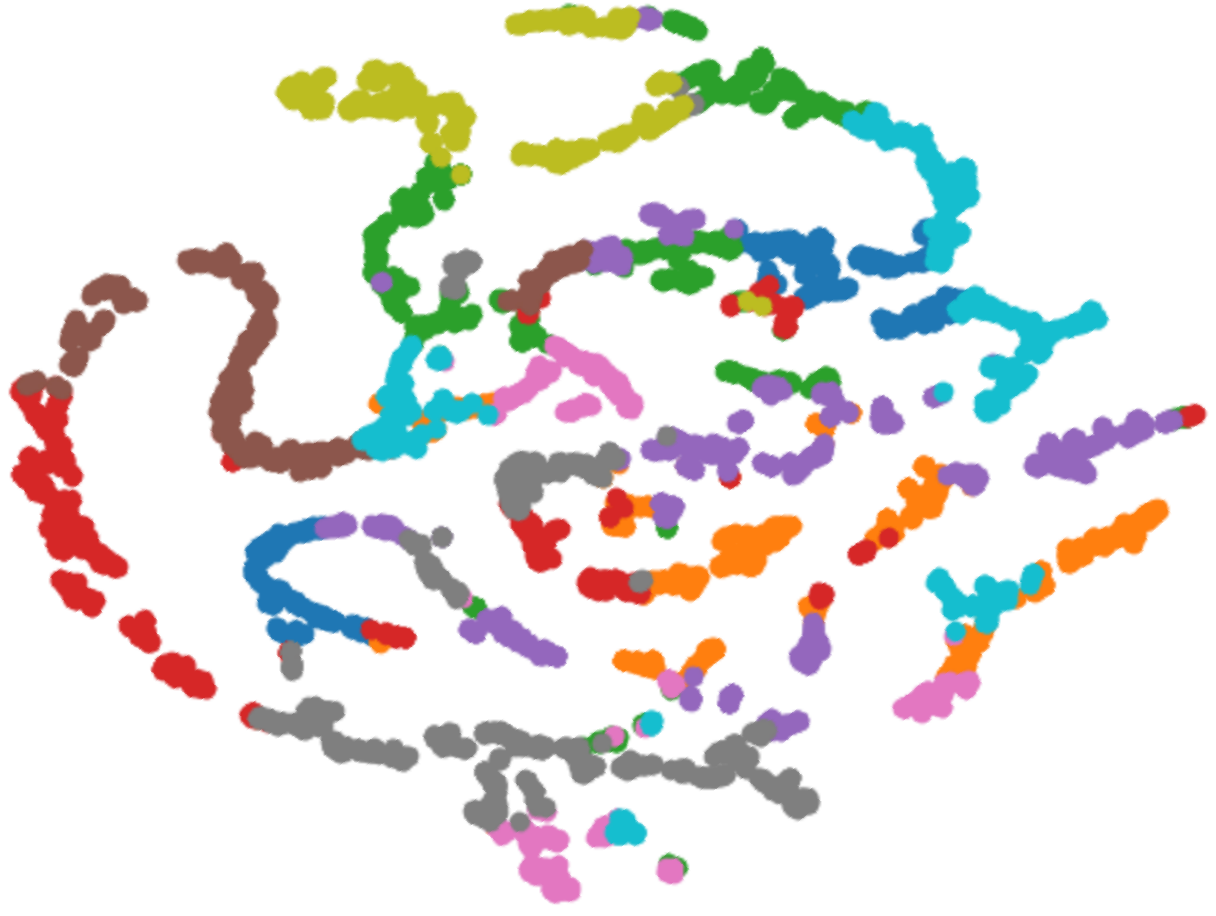}
         \vspace{-5mm}
        \caption{Fine-tuned NetVLAD}
        \label{fig:ft_netvlad_cluster}
    \end{subfigure}
     \caption{t-SNE plots color-coded by the sequence order for contrastive-based (CL) and visual place recognition (VPR)-based clustering approaches. The first three columns represent the zero-shot feature space from their respective pretrained models; the last column shows the feature space after fine-tuning on the Gibson dataset. Even in the zero-shot setting, VPR approaches exhibit a more continuous feature space, reflecting the continuous trajectory nature, compared to CL approaches. This distinction becomes even more pronounced after fine-tuning the VPR models.
     }
     \vspace{-6mm}
     \label{fig:cluster}
\end{figure*}

Imagine a mobile agent (a robot, a self-driving car, \textit{etc.}) equipped with an RGB camera while moving in an environment (either indoor or outdoor). To provide an efficient means for users to grasp essential information, scene summarization compresses a long series of input images $\mathcal{I} = { I_1, I_2, \ldots, I_n }$ into a small set of keyframes $\mathcal{K} = { K_1, K_2, \ldots, K_m }$ that capture the scene's essence, where $I_i$ and $K_i\in \mathbb{R}^{\textit{H}\times {W}\times{3}}$. Such a setup is similar to that of video summarization. However, as discussed in Sec.~\ref{sec:evaluate_metric}, scene summarization focuses on understanding the spatial scene structure and the evaluation for both tasks differs.

We propose an adaptable neural network summarizer \textbf{SceneSum} for scene summarization task, parameterized by $\phi$, to generate these keyframes without explicit frame-level importance labels. We address this task in two modes:
(1) generalization, and (2) auto-labeling. The goal of (1) refers to a model's zero-shot summarization performance on data outside the training set. In contrast, the objective of (2) refers to assessing the overfitting performance of a self-supervised model specifically on the training set.

SceneSum is a clustering-based keyframe summarizer network in two modes: (1) self-supervision, and (2) weak supervision utilizing ground truth geo-labels. The difference between these two modes is highlighted in Fig.~\ref{fig:workflow}(c). Overall, as depicted in the figure, the pipeline consists of 2 main stages and 1 optional module: (a) frames clustering, (b) keyframe selection, and (c) ground truth supervision.



\subsection{Frames Clustering}\label{sec:clustering}
Inspired by VSUMM~\cite{de2011vsumm}, clustering is used to discover patterns and group similar images within a dataset, aiding in preprocessing and organizing large datasets. In this work, we applied three ways of clustering techniques including contrastive-based, visual place recognition-based, and ground truth supervision-based techniques.

\textbf{Contrastive-based clustering.} Classic deep learning-based video summarization methods~\cite{de2011vsumm,jiang2009advances} rely on image feature for clustering, such as contrastive-based clustering methods~\cite{li2020prototypical,chen2020simple,he2020momentum} and applying K-means on ResNet/VGG features~\cite{de2011vsumm}. We applied PCL~\cite{li2020prototypical}, SimCLR~\cite{chen2020simple}, and MoCo~\cite{he2020momentum} as contrastive-based clustering baselines, because of its extensive application in visual navigation tasks in~\cite{kwon2021visual}. We present a t-SNE plot~\cite{van2008visualizing} of all contrastive-based clustering baselines from our Habitat-Sim experiment, as illustrated in Fig.~\ref{fig:MoCo_cluster}--\ref{fig:ft_pcl_cluster}. These features, though expected to be continuous as they are collected sequentially by an agent, appear relatively sparse. Besides, we further analyze the clustering performance in Sec.~\ref{sec:ab_cluster}.

\textbf{Visual place recognition based clustering.} We compared the performance of deep learning-based Visual Place Recognition (VPR) features from NetVLAD~\cite{arandjelovic2016netvlad}, Patch-NetVLAD, and MixVPR, which are widely used in SLAM tasks~\cite{garg2021seqnet,garg2022seqmatchnet,chen2023deepmapping2}. These methods aggregate local descriptors into a global representation, effectively capturing the overall \textit{visual} content and \textit{spatial} relationships within an image. The feature distribution is depicted in Fig.~\ref{fig:patch_cluster}--\ref{fig:ft_netvlad_cluster}, where we observe a continuous feature map closely resembling the actual trajectory's distribution. Notably, after fine-tuning the clustering model on our training set, this feature distribution continuity becomes even more distinct.

 \textbf{K-nearest neighbor (KNN).} KNN assigns each frame to its nearest cluster using contrastive-based and VPR-based distance metrics like Euclidean distance. For example, with $500$ frames and $k=5$, KNN partitions the images into $5$ roughly equal-sized clusters to select $5$ keyframes.

\textbf{Ground truth supervision based clustering}. We introduce a ground truth supervision module to transition our approach from self-supervised to supervised, as shown in Fig.~\ref{fig:workflow}(c). Frames are assigned to the nearest cluster based on the Euclidean distance of ground truth poses, with keyframes selected as the cluster centroids.
 
 \textbf{Cluster sampling.} Processing all images in a single batch is challenging. To enhance efficiency and robustness, we sample $N$ images from each cluster $\mathcal{C}_j$, denoted as $\mathcal{S}_j \subset \mathcal{C}_j$. This approach not only improves computational efficiency but also allows each cluster's summarized frames to \textit{derive the cluster's features from a subset of images}, similar to the way Mask Autoencoders infer features from partial data. This allows the cluster's key characteristics to be effectively captured using only a subset of images.



\subsection{Keyframe Selection}\label{sec:selection}

 After clustering stage, we focus on selecting $1$ keyframe per cluster. While classic methods like VSUMM~\cite{de2011vsumm} use the centroid of the feature space as the keyframe, this does not always correspond to the spatial centroid. Instead, we employ an encoder-decoder structure with contrastive loss and optional GT supervision to ensure that the selected keyframes are both visually and spatially separate.

\textbf{Autoencoder.}
 Autoencoders extract features that capture the essential characteristics of each image, aiding in keyframe selection through a well-represented latent space. Given an input image $\mathbf{I_i}$, an autoencoder consists of two main components: an encoder function $f(\mathbf{I_i})$ to encode image to a lower-dimensional latent space representation denoted as $\mathbf{h} = f(\mathbf{I_i})$, and a decoder function $g(\mathbf{h})$ to reconstruct the original data from the encoded representation, yielding a reconstruction $\mathbf{I_i}' = g(\mathbf{h})$.
 
 \textbf{Pooling layer}. The pooling layer reduces a 2D image feature vector ($N \times D$) of $\mathcal{S}_j$ to a 1D global cluster feature vector $p_j$ ($1 \times D$). This vector is then used for KNN search within $\mathcal{C}_j$ to identify the image with the closest feature vector to $p_j$, designating it as the keyframe for that cluster.



\subsection{Loss}
The loss function comprises two major components and one option component for ground truth supervision: the reconstruction loss, the InfoNCE loss, and the ground truth supervision loss (optional).

\textbf{Reconstruction loss.}  The reconstruction loss involves the utilization of the autoencoder for image representation learning. For each image $\mathbf{I}_i$, the autoencoder mechanism encompasses encoding and decoding stages, ultimately resulting in the reconstructed image $\mathbf{I}'_i$. The reconstruction loss $\mathcal{L}_\text { Recons.}$ is then defined as the L2 loss between the original and the reconstructed images:
\vspace{-2mm}
\begin{equation}
\mathcal{L}_\text { Recons. }=\frac{1}{N} \sum_{i=1}^N\left\|\mathbf{I}_i-\mathbf{I}_i^{\prime}\right\|_2^2
\end{equation}
\vspace{-2mm}

\textbf{InfoNCE loss.} In addition, the InfoNCE, which stands for Information Noise-Contrastive Estimation loss, plays a crucial role in separating the keyframes from each cluster by using contrastive-based loss InfoNCE. The InfoNCE principle asserts that the features of every element from any two clusters $\mathcal{S}_a$ and $\mathcal{S}_b$ should serve as \textit{negative pairs} to each other, forcing all cluster keyframes to be sufficiently distant from each other in the feature space.. Mathematically, the InfoNCE loss between clusters $\mathcal{S}_a$ and $\mathcal{S}_b$ is defined as:


\vspace{-3mm}
\begin{equation}
     \mathcal{L}_{\text{InfoNCE}_{a,b}} = -\log \frac{e^{\operatorname{sim}(\mathbf{p }_a, \mathbf{p }_a)}}{e^{\operatorname{sim}(\mathbf{p }_a, \mathbf{p }_a)}+ e^{\operatorname{sim}(\mathbf{p }_a, \mathbf{p }_b)}} 
\end{equation}
\vspace{-3mm}

where $\operatorname{sim}(.)$ is the similarity function and $e^{\operatorname{sim}(\mathbf{p }_a, \mathbf{p }_b)}$ denotes the exponential of the similarity score between  $\mathbf{p}_a$ and $\mathbf{p}_b$, $\mathbf{p}_a$ and $\mathbf{p}_b$ are the pooled(global) features of clusters $\mathcal{S}_a$ and $\mathcal{S}_b$ respectively. 

\textbf{Ground truth supervision loss.} As previously discussed in Sec.~\ref{sec:clustering}, the transition from a self-supervised to a supervised approach involves the incorporation of a ground truth (GT) supervision module. In the supervised method, apart from the change in clustering strategy, we also introduce the ground truth supervision Loss $\mathcal{L}_\text{GT}$ as:

\vspace{-2mm}
\begin{equation}
\mathcal{L}_\text{GT} = \frac{1}{N} \sum_{i=1}^{N} \| f(\mathbf{I^{GT}_i}) - \mathbf{p}_i \|_2^2 ,
\end{equation}
where $f(.)$ is the autoencoder function, $\mathbf{I^{GT}_i}$ is the keyframe selected by the ground truth pose for $i$-th cluster, $\mathbf{p}_i$ is the pooled global feature for $i$-th cluster.

\textbf{Self-supervised vs supervised loss.} For self-supervised approach, the total self-supervised loss $\mathcal{L}_{\text{self}}$ becomes the sum of the reconstruction loss and the InfoNCE loss across all pairs of clusters where

\begin{equation}
\mathcal{L}_{\text{self}}=\mathcal{L}_\text {Recons. }+\sum_{a, b\in S} \mathcal{L}_{\text{InfoNCE}_{a, b}}
\end{equation}

Besides, for supervised loss $\mathcal{L}_{\text{supervised}}$, we add $\mathcal{L}_\text{GT}$ to reduce the gap between the selected frame and the actual cluster centroid from the ground truth pose:
\begin{equation}
\mathcal{L}_{\text{supervised}}=\mathcal{L}_\text { Recons. }+\sum_{a, b\in S} \mathcal{L}_{\text{InfoNCE}_{a, b}} + \mathcal{L}_\text{GT}
\end{equation}

\vspace{-7mm}
\subsection{Evaluation Metrics}\label{sec:evaluate_metric}

 Different from the video summarization task, our work focuses on the spatial diversity of the selected keyframes. Thus, we introduce a new metric \textbf{Divergence} to evaluate the keyframe spatial diversity.

\textbf{Divergence.} We propose a new metric -- \textit{divergence} $D$ to evaluate similarity between images in a summarized set. A "similar pair" is defined as one with Euclidean distance greater than a threshold $r$. Divergence loss penalizes based on the total number of similar pairs. The diversity $D\in[0,1]$ can be summarized as:

\vspace{-1mm}
\begin{equation}
D=\frac{\sum_{I_i \in \mathcal{K}}\left|d_i\right|}{|\mathcal{K}|^2},
\end{equation}
\vspace{-2.2mm}
\begin{equation}
d_i=\left\{|I_j|\mid| T(I_i)-T(I_j) \mid<r, I_j \in \mathcal{K}\right\}
\end{equation}

$I_i$ and $I_j$ represent individual images in the keyframe set $\mathcal{K}$, $T(.)$  represents the odometry position where the image is taken, $d_i$ is the total number of similar pairs for keyframe $i$, $r$ is a distance threshold in meters. Methods with lower divergence (\textit{D}) at the same distance threshold (\textit{r}) are deemed superior, indicating that each keyframe has fewer neighboring keyframes on average.

\textbf{Area Under the Curve (AUC) on divergence
.} AUC measures the area under the divergence curve, which plots \textit{diversity} against distance threshold \textit{r}. It reflects the average number of neighboring keyframes across thresholds.

%% file: sections/4-experiments.tex
\section{Experiment}\label{sec:experiment}
\vspace{-1mm}

\begin{figure*}[t]
  \centering
  \setlength{\abovecaptionskip}{4pt}
  \begin{minipage}{1\textwidth} 
    \includegraphics[width=\linewidth]{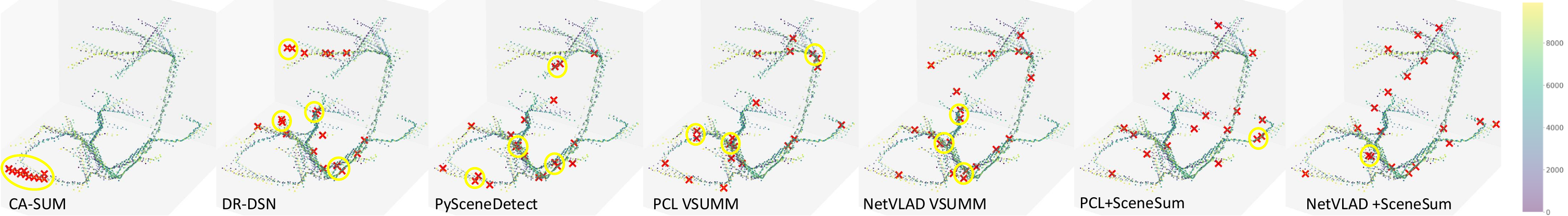}
    \caption{\textbf{Selected keyframes in Habitat-Sim Dataset.} We summarize $20$ keyframes of $7$ baselines on scene \textit{Stokes}. All frames are color-coded by temporal order. Summarized keyframes are marked with red crosses. Groups of frames that are geographically close to each other are circled in yellow.}
    \label{fig:habitat_visualization}
  \end{minipage}\hfill
\end{figure*}

\begin{figure*}[t]
  \centering
  \setlength{\abovecaptionskip}{4pt}
  \begin{minipage}{1\textwidth} 
    \includegraphics[width=\linewidth]{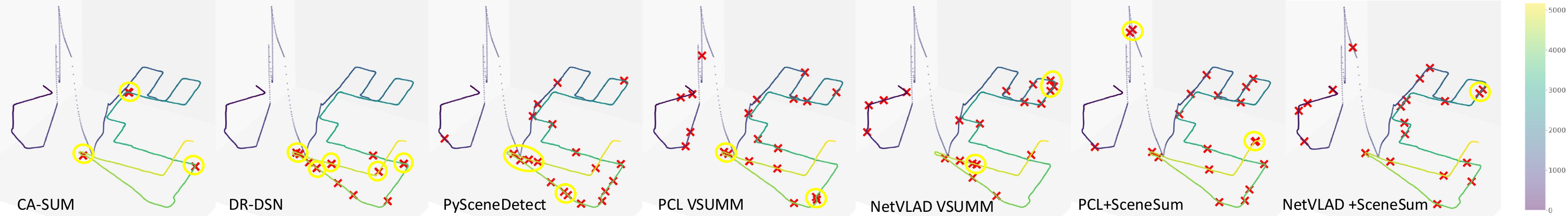}
    \caption{\textbf{Selected keyframes in KITTI Dataset.} We summarize $20$ keyframes of $7$ baselines on scene \textit{0028}. The baselines and annotations follow Fig.~\ref{fig:habitat_visualization}}
    \label{fig:kitti_visualization}
  \end{minipage}\hfill
\vspace{-4mm}
\end{figure*}

 \textbf{Overview.} This section focuses on evaluating the effectiveness and adaptability of SceneSum in the realm of scene summarization. We conduct various analyses on two different datasets: the ``Habitat-Sim" -- simulated indoor panoramic RGB dataset (Sec.~\ref{sec:habitat_experiment}) and the ``KITTI" -- real-world outdoor perspective RGB dataset (Sec.~\ref{sec:kitti_experiment}). We evaluate SceneSum against prominent video summarization methods like PySceneDetect~\cite{pyscenedetect2017}, DR-DSN~\cite{zhou2018deep}, CA-SUM~\cite{apostolidis2022summarizing} , and VSUMM~\cite{de2011vsumm} by divergence and divergence AUC detailed in Sec.~\ref{sec:evaluate_metric}.

Besides, we also conduct extensive ablation research \textit{in the appendix}, delving into a comparative analysis of (1) supervised vs self-supervised method, (2) divergence comparison on different baselines, and (3) auto-labeling capability. 

\vspace{-2mm}
\subsection{Experiment Settings}\label{sec:Settings}
\vspace{-1mm}
\textbf{Datasets}\label{Datasets}.
We evaluate SceneSum on a
simulated RGB dataset using the Habitat-Sim simulator\cite{savva2019habitat}
based on the Gibson photorealistic RGB dataset\cite{xiazamirhe2018gibsonenv}.
This dataset includes panoramic RGB images captured from various indoor scans. A panoramic camera mounted on a robot navigated randomly through the virtual environments, resulting in $72,792$ RGB images, each downsampled to $512\times 256$.

Besides, we also conducted evaluations of SceneSum on outdoor non-panoramic RGB datasets - KITTI~\cite{kitti}. This dataset is recorded by the Point Grey Flea 2 (FL2-14S3C-C) sensor, configured in a roof-mounted pushbroom setup. This dataset comprises $12,483$ RGB images covering a driving distance of $73.7$ kilometers. After pre-processing, we resize each frame to
an image of size $512\times 256$, with each image corresponding to the exact GPS location. Our evaluation focuses on the three most complex scenes within the KITTI dataset, where multiple revisits to various locations occur, and the explored area is considerably vast.

\begin{table*}[h]
\centering
\setlength{\abovecaptionskip}{3pt}
\caption{Best AUC results for the Habitat-Sim dataset. We conducted experiments under diverse cluster sizes across different scenes, reporting AUC, average (AVG), and standard deviation (SD) values. We emphasize the top three performing baselines using \textcolor{red}{red}, \textcolor{teal}{teal}, and \textcolor{blue}{blue} highlights respectively, with lower values indicating superior performance.}
\label{tab:habitat}
\resizebox{1\linewidth}{!}{
\begin{tabular}{|c|c|c|c|c|c|c|c|c|c|c|c|c|c|c|c|c|c|c|}
\hline
\multirow{2}{*}{\begin{tabular}[c]{@{}c@{}}Scene\\ Number of Clusters\end{tabular}} & \multicolumn{6}{c|}{\textbf{Micanopy(↓)}} & \multicolumn{6}{c|}{\textbf{Spotswood(↓)}} & \multicolumn{6}{c|}{\textbf{Springhill(↓)}} \\ \cline{2-19}
 & \textbf{10} & \textbf{20} & \textbf{30} & \textbf{40} & \textbf{AVG.} & \textbf{SD.} & \textbf{10} & \textbf{20} & \textbf{30} & \textbf{40}  & \textbf{AVG.} & \textbf{SD.} & \textbf{10} & \textbf{20} & \textbf{30} & \textbf{40} & \textbf{AVG.} & \textbf{SD.}\\ \hline
 CA-SUM& \textcolor{black}{0.309} & \textcolor{black}{0.252} & \textcolor{black}{0.254} & \textcolor{black}{0.241} & \textcolor{black}{0.264} &\textcolor{black}{0.031} & \textcolor{teal}{0.082}& \textcolor{blue}{0.195}& \textcolor{blue}{0.206}& \textcolor{black}{0.288} & \textcolor{blue}{0.193}  &\textcolor{black}{0.085}& \textcolor{black}{0.585} & \textcolor{black}{0.487} & \textcolor{black}{0.357} & \textcolor{black}{0.301} & \textcolor{black}{0.433} &\textcolor{black}{0.128}\\ 
DR-DSN & \textcolor{red}{0.045}& \textcolor{blue}{0.172}& \textcolor{blue}{0.180}& \textcolor{black}{0.205}& \textcolor{teal}{0.151} &\textcolor{black}{0.072} & \textcolor{black}{0.345} & \textcolor{black}{0.457} & \textcolor{black}{0.316} & \textcolor{black}{0.291}& \textcolor{black}{0.352} &\textcolor{black}{0.073} & \textcolor{black}{0.105}& \textcolor{black}{0.240} & \textcolor{black}{0.273} & \textcolor{black}{0.290}& \textcolor{black}{0.227} &\textcolor{black}{0.084}\\ 
\textcolor{black}{IPTNET} & \textcolor{black}{0.533}& \textcolor{black}{0.522}& \textcolor{black}{0.402} & \textcolor{black}{0.374}& \textcolor{black}{0.458} & \textcolor{black}{0.081} & \textcolor{black}{0.317}& \textcolor{black}{0.405} & \textcolor{black}{0.359} & \textcolor{black}{0.306}& \textcolor{black}{0.347} & \textcolor{black}{0.045} & \textcolor{black}{0.471}& \textcolor{black}{0.244}& \textcolor{black}{0.243} & \textcolor{black}{0.215}& \textcolor{blue}{0.293}& \textcolor{black}{0.119} 
\\
PCL VSUMM & \textcolor{black}{0.158}& \textcolor{black}{0.218} & \textcolor{black}{0.257}& \textcolor{black}{0.250}& \textcolor{black}{0.221} & \textcolor{black}{0.045} & \textcolor{black}{0.187} & \textcolor{black}{0.234} & \textcolor{black}{0.271} & \textcolor{black}{0.230}& \textcolor{black}{0.230} & \textcolor{black}{0.034} & \textcolor{teal}{0.080}& \textcolor{black}{0.179} & \textcolor{blue}{0.202}& \textcolor{black}{0.219}& \textcolor{blue}{0.170} & \textcolor{black}{0.062} \\ 
NetVLAD VSUMM & \textcolor{black}{0.254}& \textcolor{black}{0.324} & \textcolor{black}{0.312}& \textcolor{black}{0.332}& \textcolor{black}{0.306}&  \textcolor{black}{0.035} &  \textcolor{black}{0.229} & \textcolor{black}{0.321} & \textcolor{black}{0.224} & \textcolor{black}{0.237}& \textcolor{black}{0.253} &  \textcolor{black}{0.046} &\textcolor{black}{0.183}& \textcolor{blue}{0.141} & \textcolor{black}{0.238}& \textcolor{blue}{0.178}& \textcolor{black}{0.185} & \textcolor{black}{0.040} \\ 
PySceneDetect & \textcolor{black}{0.157}& \textcolor{black}{0.197} & \textcolor{teal}{0.172}& \textcolor{blue}{0.184}& \textcolor{black}{0.178} &\textcolor{black}{0.017} &\textcolor{black}{0.358} & \textcolor{black}{0.218} & \textcolor{black}{0.218} & \textcolor{teal}{0.174}& \textcolor{black}{0.242} &\textcolor{black}{0.080}&  \textcolor{black}{0.108} & \textcolor{black}{0.197}& \textcolor{black}{0.207}& \textcolor{black}{0.224}& \textcolor{black}{0.184} &\textcolor{black}{0.052}\\ 
\hline
PCL+SceneSum & \textcolor{blue}{0.155}& \textcolor{teal}{0.152}& \textcolor{black}{0.184}& \textcolor{teal}{0.172} & \textcolor{blue}{0.166} & \textcolor{black}{0.015} & \textcolor{blue}{0.117}& \textcolor{teal}{0.151}& \textcolor{red}{0.146}& \textcolor{blue}{0.215}& \textcolor{teal}{0.171} & \textcolor{black}{0.041} & \textcolor{red}{0.023}& \textcolor{red}{0.086}& \textcolor{red}{0.129}& \textcolor{red}{0.106}& \textcolor{red}{0.086} & \textcolor{black}{0.046} \\ 
NetVLAD+SceneSum & \textcolor{teal}{0.118} & \textcolor{red}{0.117}& \textcolor{red}{0.167}& \textcolor{red}{0.159}& \textcolor{red}{0.140} & \textcolor{black}{0.026} & \textcolor{red}{0.067}& \textcolor{red}{0.120}& \textcolor{teal}{0.191}& \textcolor{red}{0.163}& \textcolor{red}{0.135}& \textcolor{black}{0.054} & \textcolor{blue}{0.101}& \textcolor{teal}{0.111}& \textcolor{teal}{0.140}& \textcolor{teal}{0.138}& \textcolor{teal}{0.123}& \textcolor{black}{0.020} \\ 
\hline

\hline
\hline

\multirow{2}{*}{\begin{tabular}[c]{@{}c@{}}Scene\\ Number of Clusters\end{tabular}} & \multicolumn{6}{c|}{\textbf{Stilwell(↓)}} & \multicolumn{6}{c|}{\textbf{Stokes(↓)}} & \multicolumn{6}{c|}{\textbf{Goffs(↓)}} \\ \cline{2-19}
 & \textbf{10} & \textbf{20} & \textbf{30} & \textbf{40} & \textbf{AVG.} & \textbf{SD.} & \textbf{10} & \textbf{20} & \textbf{30} & \textbf{40}  & \textbf{AVG.} & \textbf{SD.} & \textbf{10} & \textbf{20} & \textbf{30} & \textbf{40}  & \textbf{AVG.} & \textbf{SD.} \\ \hline
\textcolor{black}{CA-SUM} & \textcolor{black}{0.408} & \textcolor{black}{0.543} & \textcolor{black}{0.424} & \textcolor{black}{0.370}& \textcolor{black}{0.436} & \textcolor{black}{0.075} & \textcolor{black}{1.968} & \textcolor{black}{1.791} & \textcolor{black}{0.915} & \textcolor{black}{0.738}& \textcolor{black}{1.353} & \textcolor{black}{0.616} & \textcolor{black}{0.293} & \textcolor{black}{0.190} & \textcolor{black}{0.176} & \textcolor{black}{0.195}& \textcolor{black}{0.214}& \textcolor{black}{0.054} 
\\ 
\textcolor{black}{DR-DSN} & \textcolor{black}{0.176}& \textcolor{black}{0.225}& \textcolor{black}{0.214} & \textcolor{black}{0.222}& \textcolor{black}{0.209} & \textcolor{black}{0.023} & \textcolor{black}{0.206}& \textcolor{black}{0.287} & \textcolor{black}{0.308} & \textcolor{black}{0.344}& \textcolor{black}{0.286} & \textcolor{black}{0.058} & \textcolor{teal}{0.059}& \textcolor{blue}{0.118}& \textcolor{black}{0.182} & \textcolor{black}{0.181}& \textcolor{blue}{0.135}& \textcolor{black}{0.059} 
\\
\textcolor{black}{IPTNET} & \textcolor{black}{0.291}& \textcolor{black}{0.245}& \textcolor{black}{0.325} & \textcolor{black}{0.269}& \textcolor{black}{0.283} & \textcolor{black}{0.034} & \textcolor{black}{0.359}& \textcolor{black}{0.273} & \textcolor{black}{0.264} & \textcolor{black}{0.237}& \textcolor{black}{0.286} & \textcolor{black}{0.053} & \textcolor{black}{0.237}& \textcolor{black}{0.293}& \textcolor{black}{0.235} & \textcolor{black}{0.210}& \textcolor{blue}{0.259}& \textcolor{black}{0.043} 
\\
PCL VSUMM & \textcolor{black}{0.272} & \textcolor{teal}{0.132} & \textcolor{black}{0.235} & \textcolor{black}{0.214}& \textcolor{black}{0.213}& \textcolor{black}{0.059} & \textcolor{blue}{0.117}& \textcolor{black}{0.227} & \textcolor{black}{0.290}& \textcolor{black}{0.248}& \textcolor{black}{0.221} & \textcolor{black}{0.074} & \textcolor{black}{0.104} & \textcolor{black}{0.163} & \textcolor{black}{0.160} & \textcolor{black}{0.173}& \textcolor{black}{0.150} & \textcolor{black}{0.031} \\ 
NetVLAD VSUMM & \textcolor{blue}{0.107} & \textcolor{black}{0.210} & \textcolor{black}{0.262} & \textcolor{black}{0.250}& \textcolor{black}{0.207} & \textcolor{black}{0.070}  & \textcolor{black}{0.267}& \textcolor{blue}{0.213} & \textcolor{black}{0.293}& \textcolor{black}{0.290}& \textcolor{black}{0.266} & \textcolor{black}{0.037} & \textcolor{black}{0.080} & \textcolor{black}{0.205} & \textcolor{blue}{0.147} & \textcolor{black}{0.183}& \textcolor{black}{0.154}& \textcolor{black}{0.055} \\
\textcolor{black}{PySceneDetect} & \textcolor{black}{0.134}& \textcolor{black}{0.260} & \textcolor{blue}{0.174}& \textcolor{blue}{0.183}& \textcolor{blue}{0.188} & \textcolor{black}{0.053} & \textcolor{black}{0.246} & \textcolor{black}{0.235}& \textcolor{blue}{0.161}& \textcolor{blue}{0.202}& \textcolor{blue}{0.211}& \textcolor{black}{0.038} & \textcolor{black}{0.091}& \textcolor{black}{0.181} & \textcolor{black}{0.155}& \textcolor{blue}{0.151}& \textcolor{black}{0.145} & \textcolor{black}{0.038} \\ 
\hline
\textcolor{black}{PCL+SceneSum} & \textcolor{teal}{0.093} & \textcolor{blue}{0.162}& \textcolor{teal}{0.154}& \textcolor{teal}{0.147}& \textcolor{teal}{0.139} & \textcolor{black}{0.031} & \textcolor{teal}{0.094}& \textcolor{red}{0.108}& \textcolor{red}{0.157}& \textcolor{teal}{0.193}& \textcolor{teal}{0.138}& \textcolor{black}{0.046} & \textcolor{blue}{0.073}& \textcolor{red}{0.086}& \textcolor{teal}{0.107}& \textcolor{red}{0.112}& \textcolor{teal}{0.095}& \textcolor{black}{0.018} \\ 
\textcolor{black}{NetVLAD+SceneSum} & \textcolor{red}{0.039}& \textcolor{red}{0.091}& \textcolor{red}{0.130}& \textcolor{red}{0.132}& \textcolor{red}{0.103}& \textcolor{black}{0.044} & \textcolor{red}{0.076}& \textcolor{teal}{0.109}& \textcolor{teal}{0.158}& \textcolor{red}{0.154}& \textcolor{red}{0.124}& \textcolor{black}{0.039} & \textcolor{red}{0.050}& \textcolor{teal}{0.106}& \textcolor{red}{0.096}& \textcolor{teal}{0.119}& \textcolor{red}{0.093}& \textcolor{black}{0.030} \\ 
\hline

\hline

\hline
\hline
\multirow{2}{*}{\begin{tabular}[c]{@{}c@{}}Scene\\ Number of Clusters\end{tabular}} & \multicolumn{18}{c|}{\textbf{Average Result(↓)}} 
\\ \cline{2-19}
 & \multicolumn{3}{c|}{\textbf{10}} & \multicolumn{3}{c|}{\textbf{20}} & \multicolumn{3}{c|}{\textbf{30}} & \multicolumn{3}{c|}{\textbf{40}} & \multicolumn{3}{c|}{\textbf{AVG.}}  & \multicolumn{3}{c|}{\textbf{SD.}}\\ \hline
 \textcolor{black}{CA-SUM} & \multicolumn3{c|}{\textcolor{black}{0.608}} & \multicolumn{3}{c|}{\textcolor{black}{0.576}} & \multicolumn{3}{c|}{\textcolor{black}{0.389}} & \multicolumn{3}{c|}{\textcolor{black}{0.356}} & \multicolumn{3}{c|}{\textcolor{black}{0.482}} & \multicolumn{3}{c|}{\textcolor{black}{0.128}}\\
\textcolor{black}{DR-DSN} & \multicolumn{3}{c|}{\textcolor{black}{0.156}} & \multicolumn{3}{c|}{\textcolor{black}{0.250}} & \multicolumn{3}{c|}{\textcolor{black}{0.246}} & \multicolumn{3}{c|}{\textcolor{black}{0.256}} & \multicolumn{3}{c|}{\textcolor{black}{0.227}}& \multicolumn{3}{c|}{\textcolor{black}{0.048}} \\
\textcolor{black}{IPTNET} & \multicolumn{3}{c|}{\textcolor{black}{0.378}} & \multicolumn{3}{c|}{\textcolor{black}{0.330}} & \multicolumn{3}{c|}{\textcolor{black}{0.305}} & \multicolumn{3}{c|}{\textcolor{black}{0.269}} & \multicolumn{3}{c|}{\textcolor{black}{0.321}}& \multicolumn{3}{c|}{\textcolor{black}{0.046}} \\
\textcolor{black}{PCL VSUMM} & \multicolumn{3}{c|}{\textcolor{blue}{0.153}} & \multicolumn{3}{c|}{\textcolor{blue}{0.192}} & \multicolumn{3}{c|}{\textcolor{black}{0.236}} & \multicolumn{3}{c|}{\textcolor{black}{0.222}} & \multicolumn{3}{c|}{\textcolor{black}{0.201}}& \multicolumn{3}{c|}{\textcolor{black}{0.037}}\\
\textcolor{black}{NetVLAD VSUMM} & \multicolumn{3}{c|}{\textcolor{black}{0.187}} & \multicolumn{3}{c|}{\textcolor{black}{0.236}} & \multicolumn{3}{c|}{\textcolor{black}{0.246}} & \multicolumn{3}{c|}{\textcolor{black}{0.246}} & \multicolumn{3}{c|}{\textcolor{black}{0.229}}& \multicolumn{3}{c|}{\textcolor{black}{0.028}}\\
\textcolor{black}{PySceneDetect} & \multicolumn{3}{c|}{\textcolor{black}{0.182}} & \multicolumn3{c|}{\textcolor{black}{0.215}} & \multicolumn{3}{c|}{\textcolor{blue}{0.181}} & \multicolumn{3}{c|}{\textcolor{blue}{0.186}} & \multicolumn{3}{c|}{\textcolor{blue}{0.191}}& \multicolumn{3}{c|}{\textcolor{black}{0.016}}\\
\hline
\textcolor{black}{PCL+SceneSum} & \multicolumn{3}{c|}{\textcolor{teal}{0.093}} & \multicolumn{3}{c|}{\textcolor{teal}{0.124}} & \multicolumn{3}{c|}{\textcolor{red}{0.146}} & \multicolumn{3}{c|}{\textcolor{teal}{0.158}} & \multicolumn{3}{c|}{\textcolor{teal}{0.133}}& \multicolumn{3}{c|}{\textcolor{black}{0.029}}\\
\textcolor{black}{NetVLAD+SceneSum} & \multicolumn{3}{c|}{\textcolor{red}{0.075}} & \multicolumn{3}{c|}{\textcolor{red}{0.113}} & \multicolumn{3}{c|}{\textcolor{teal}{0.147}} & \multicolumn{3}{c|}{\textcolor{red}{0.144}} & \multicolumn{3}{c|}{\textcolor{red}{0.120}}& \multicolumn{3}{c|}{\textcolor{black}{0.034}}\\
\hline

\end{tabular}
}
\vspace{-2mm}
\end{table*}

\textbf{Baseline Methods.} 
The following methods are compared to highlight the effectiveness of the proposed SceneSum.
(1) \textit{PySceneDetect}. A classic non-learning approach based on color histograms~\cite{pyscenedetect2017}, (2) \textit{DR-DSN} ~\cite{zhou2018deep}, and (3) \textit{CA-SUM.}~\cite{apostolidis2022summarizing} -- A deep learning-based summarizers, (4) \textit{VSUMM} -- A clustering-based video summarizer with contrastive-based and VPR-based clustering~\cite{de2011vsumm}, 
(5) \textit{SceneSum}. The proposed framework of clustering for summarizing scene video. Our framework has three variants for evaluation: SceneSum with contrastive-based clustering, SceneSum with VPR-based clustering, and ground truth supervised SceneSum. We use a batch size of $64$, a learning rate of $0.001$, and an embedding dimension of $2048$. The model is trained with the Adam optimizer for $100$ epochs.

\vspace{-1mm}
\subsection{More ablations on clustering}\label{sec:ab_cluster}
\vspace{-1mm}
In Sec.~\ref{sec:clustering}, we visualized the continuity in the feature space when comparing contrastive-based clustering and VPR clustering. To further demonstrate the relevance of this feature space continuity to scene summarization, we conducted an ablation study: using six different baselines for clustering in various Habitat-Sim scenes, combined with our keyframe selection
 stage. Then we obtained the average values for the divergence AUC shown in Tab.~\ref{tab:div_ablations}. Compared to the VPR-based clustering approach, traditional contrastive-based clustering methods for summarization are less effective due to their limited ability to capture spatial context. The contrastive features typically encode spatial relations in a local context but do not naturally extend this encoding to a global spatial context that is vital for understanding the entire scene's layout. Therefore, we selected the best contrastive method, PCL, and the worst VPR method, NetVLAD, for the subsequent experiments.

\begin{table}[h!]
\centering
\vspace{-1mm}
\resizebox{0.48\textwidth}{!}{%
\begin{tabular}{|c|c|c|c|c|c|c|}
\hline
\textbf{Method} & MoCo & SimCLR & PCL & NetVLAD & Patch-NetVLAD & MixVPR \\
\hline
\textbf{Div. AUC (↓)} & 0.3397 & 0.4259 & 0.133 & 0.120 & \textbf{0.083} & 0.094 \\
\hline
\end{tabular}%
}
\vspace{-3mm}
\caption{Comparison of the average divergence AUC across six methods in various scenes from the Habitat-Sim dataset.}
\label{tab:div_ablations}
\vspace{-4mm}
\end{table}

\vspace{-1mm}
\subsection{Habitat-Sim Experiment}\label{sec:habitat_experiment}
\vspace{-1mm}

\textbf{Implementation details.}
We investigate multiple hyperparameters including the distance threshold $r \in [0, 3]$ meters for Habitat-Sim and the size of the targeted keyframe set $k \in \{10, 20, 30, 40\}$. Besides, we selected $6$ scenes from the Habitat-Sim dataset for training and testing. We conduct $6$ cross-validation experiments on the Habitat-Sim dataset, wherein each experiment involves training the model on five scenes for 100 epochs utilizing the NVIDIA A100 GPU, followed by inference on the remaining scene. In this section, we only discuss self-supervised SceneSum.

\textbf{Baseline comparisons.}
In Tab.~\ref{tab:habitat}, SceneSum is compared with several video summarization methods. Among all, CA-SUM underperforms, highlighting its limitations in selecting spatially diverse keyframes. Moreover, PCL VSUMM and NetVLAD VSUMM demonstrated nearly identical, enhanced results in comparison to CA-SUM. Notably, DR-DSN performs well when using fewer clusters. Lastly, PySceneDetect emerged as the most effective among all the baselines. 

Most significantly, SceneSum outperforms these baselines in terms of the Area Under the Curve (AUC) metric across various cluster sizes, affirming its ability to select more spatially diverse keyframes.  Analysis of the data from Tab.\ref{tab:habitat} reveals that SceneSum significantly outperforms all baselines across different Habitat-Sim scenes. This superiority of SceneSum is further evidenced in Fig.\ref{fig:habitat_visualization}, where it typically showcases a more dispersed frame distribution, validating our two-stage model's efficacy in scene summarization.
Improvements were noted across regions and cluster sizes: Micanopy saw a 7.86\% increase, Spotswood 42.9\%, Springhill 97.6\%, Stilwell 82.5\%, Stokes 70.1\%, and Goffs 45.2\%. In cluster analysis, increases were 104\% for 10 clusters, 69.9\% at 20, 24.0\% at 30, and 29.2\% at 40.

\vspace{-1mm}
\subsection{KITTI Experiment}\label{sec:kitti_experiment}
\vspace{-1mm}
\begin{table*}[h]
\centering
\setlength{\abovecaptionskip}{3pt}
\caption{Best Divergence AUC for KITTI dataset. We do experiments across three scenes. Other setups and abbreviations follow Tab.~\ref{tab:habitat}}
\label{tab:kitti}
\resizebox{0.98\linewidth}{!}{
\begin{tabular}{|c|c|c|c|c|c|c|c|c|c|c|c|c|}
\hline
\multirow{2}{*}{\begin{tabular}[c]{@{}c@{}}Scene\\ Number of Clusters\end{tabular}} & \multicolumn{6}{c|}{\textbf{KITTI (0018)(↓)}} & \multicolumn{6}{c|}{\textbf{KITTI (0027)(↓)}} \\ \cline{2-13}
 & \textbf{10} & \textbf{20} & \textbf{30} & \textbf{40} & \textbf{AVG.} & \textbf{SD.}& \textbf{10} & \textbf{20} & \textbf{30} & \textbf{40}  & \textbf{AVG.}& \textbf{SD.} \\ \hline

CA-SUM& \textcolor{black}{98.155} & \textcolor{black}{69.287} & \textcolor{black}{52.288} & \textcolor{black}{42.290}& \textcolor{black}{65.505}& \textcolor{black}{24.454} & \textcolor{black}{49.733}& \textcolor{black}{47.028}& \textcolor{black}{50.103}& \textcolor{black}{44.141}& \textcolor{black}{47.751} &\textcolor{black}{2.77}\\ 
 
DR-DSN & \textcolor{black}{95.678}& \textcolor{black}{46.136}& \textcolor{black}{23.504}& \textcolor{black}{24.485}& \textcolor{black}{47.450}& \textcolor{black}{33.806}& \textcolor{black}{17.777} & \textcolor{black}{14.442} & \textcolor{black}{13.139} & \textcolor{black}{14.793}& \textcolor{black}{15.037}& \textcolor{black}{1.960} \\ 
\textcolor{black}{IPTNET} & \textcolor{black}{96.433}& \textcolor{black}{30.242} & \textcolor{black}{32.715} & \textcolor{black}{24.079}& \textcolor{black}{45.867}& \textcolor{black}{33.906} & \textcolor{black}{34.978}& \textcolor{black}{16.253}& \textcolor{black}{14.100} & \textcolor{black}{19.887}& \textcolor{black}{21.305}& \textcolor{black}{9.423}\\ 

PCL VSUMM & \textcolor{black}{11.155}& \textcolor{black}{7.297} & \textcolor{teal}{6.564}& \textcolor{black}{10.566}& \textcolor{black}{8.895}&\textcolor{black}{2.301}& \textcolor{black}{4.288} & \textcolor{blue}{3.502} & \textcolor{blue}{3.409} & \textcolor{teal}{4.562}& \textcolor{black}{3.940}&\textcolor{black}{0.572}\\ 

NetVLAD VSUMM & \textcolor{black}{9.166}& \textcolor{black}{7.85} & \textcolor{black}{8.394}& \textcolor{black}{9.591}& \textcolor{black}{8.750} & \textcolor{black}{0.778} & \textcolor{black}{4.355} & \textcolor{black}{5.076} & \textcolor{black}{6.414} & \textcolor{black}{6.037}& \textcolor{black}{5.470} & \textcolor{black}{0.933}\\

PySceneDetect & \textcolor{teal}{5.411}& \textcolor{teal}{6.015} & \textcolor{blue}{6.700}& \textcolor{blue}{8.113}& \textcolor{teal}{6.559}&\textcolor{black}{1.162}& \textcolor{blue}{2.911} & \textcolor{teal}{2.939} & \textcolor{black}{3.596} & \textcolor{blue}{4.691}& \textcolor{blue}{3.534} &\textcolor{black}{0.834}\\ 

\hline

PCL+SceneSum & \textcolor{blue}{8.955}& \textcolor{blue}{6.078}& \textcolor{black}{7.181}& \textcolor{teal}{6.866}& \textcolor{blue}{7.270}& \textcolor{black}{1.215} & \textcolor{teal}{0.655}& \textcolor{black}{4.434}& \textcolor{red}{2.980}& \textcolor{black}{4.717}& \textcolor{teal}{3.196}& \textcolor{black}{1.857}\\ 
NetVLAD+SceneSum & \textcolor{red}{2.533} & \textcolor{red}{4.703}& \textcolor{red}{5.688}& \textcolor{red}{6.606}& \textcolor{red}{4.883}& \textcolor{black}{1.748}& \textcolor{red}{0.456}& \textcolor{red}{1.626}& \textcolor{teal}{3.094}& \textcolor{red}{3.831}& \textcolor{red}{2.623}& \textcolor{black}{1.508}\\ 

\hline
\hline

\multirow{2}{*}{\begin{tabular}[c]{@{}c@{}}Scene\\ Number of Clusters\end{tabular}} & \multicolumn{6}{c|}{\textbf{KITTI (0028)(↓)}} & \multicolumn{6}{c|}{\textbf{Average Result(↓)}} \\ \cline{2-13}
 & \textbf{10} & \textbf{20} & \textbf{30} & \textbf{40} & \textbf{AVG.} & \textbf{SD.} & \textbf{10} & \textbf{20} & \textbf{30} & \textbf{40}  & \textbf{AVG.} & \textbf{SD.} \\ \hline

\textcolor{black}{CA-SUM} & \textcolor{black}{20.922} & \textcolor{black}{56.407} & \textcolor{black}{41.347} & \textcolor{black}{31.895} & \textcolor{black}{37.642} & \textcolor{black}{15.038}& \textcolor{black}{56.270} & \textcolor{black}{57.574} & \textcolor{black}{47.913} & \textcolor{black}{39.442}& \textcolor{black}{50.300}& \textcolor{black}{8.409}\\ 

\textcolor{black}{DR-DSN} & \textcolor{black}{10.399}& \textcolor{black}{10.915} & \textcolor{black}{11.565} & \textcolor{black}{12.444}& \textcolor{black}{11.330}& \textcolor{black}{0.882} & \textcolor{black}{41.285}& \textcolor{black}{23.831}& \textcolor{black}{16.069} & \textcolor{black}{17.241}& \textcolor{black}{24.606}& \textcolor{black}{11.632}\\ 
\textcolor{black}{IPTNET} & \textcolor{black}{19.567}& \textcolor{black}{17.211} & \textcolor{black}{22.513} & \textcolor{black}{20.769}& \textcolor{black}{20.015}& \textcolor{black}{2.226} & \textcolor{black}{50.326}& \textcolor{black}{21.235}& \textcolor{black}{23.109} & \textcolor{black}{21.578}& \textcolor{black}{29.062}& \textcolor{black}{14.199}\\ 

PCL VSUMM & \textcolor{blue}{0.388}& \textcolor{blue}{2.342} & \textcolor{blue}{2.951}& \textcolor{black}{3.056}& \textcolor{blue}{2.184}& \textcolor{black}{1.238}& \textcolor{black}{5.277} & \textcolor{black}{4.380} & \textcolor{blue}{4.308} & \textcolor{black}{6.061}& \textcolor{black}{5.007}& \textcolor{black}{0.830}\\ 
NetVLAD VSUMM & \textcolor{black}{3.177}& \textcolor{black}{3.305} & \textcolor{black}{3.235}& \textcolor{blue}{3.039}& \textcolor{black}{3.189}& \textcolor{black}{0.113}& \textcolor{black}{5.566} & \textcolor{black}{5.410} & \textcolor{black}{6.014} & \textcolor{black}{6.222}& \textcolor{black}{5.803}& \textcolor{black}{0.379}\\ 
\textcolor{black}{PySceneDetect} & \textcolor{black}{0.766} & \textcolor{black}{3.539}& \textcolor{black}{3.194}& \textcolor{black}{3.660}& \textcolor{black}{2.789}& \textcolor{black}{1.364}& \textcolor{teal}{3.029}& \textcolor{blue}{4.164} & \textcolor{black}{4.496}& \textcolor{blue}{5.488}& \textcolor{blue}{4.294}& \textcolor{black}{1.014}\\
\hline
\textcolor{black}{PCL+SceneSum} & \textcolor{teal}{0.005}& \textcolor{teal}{1.736}& \textcolor{teal}{2.650}& \textcolor{teal}{2.508}& \textcolor{teal}{1.926}& \textcolor{black}{1.215}& \textcolor{blue}{3.205} & \textcolor{teal}{4.083}& \textcolor{teal}{4.270}& \textcolor{teal}{4.697}& \textcolor{teal}{4.064}& \textcolor{black}{0.628}\\

\textcolor{black}{NetVLAD+SceneSum} & \textcolor{red}{0.000}& \textcolor{red}{1.231}& \textcolor{red}{1.480}& \textcolor{red}{2.321}& \textcolor{red}{1.391}& \textcolor{black}{0.960}& \textcolor{red}{0.996}& \textcolor{red}{2.520}& \textcolor{red}{3.421}& \textcolor{red}{4.253}& \textcolor{red}{2.798}& \textcolor{black}{1.394}\\ 


\hline

\end{tabular}}
\vspace{-2mm}
\end{table*}

\begin{figure*}[h!]
\vspace{1mm}
    \centering {\includegraphics[width=1\textwidth]{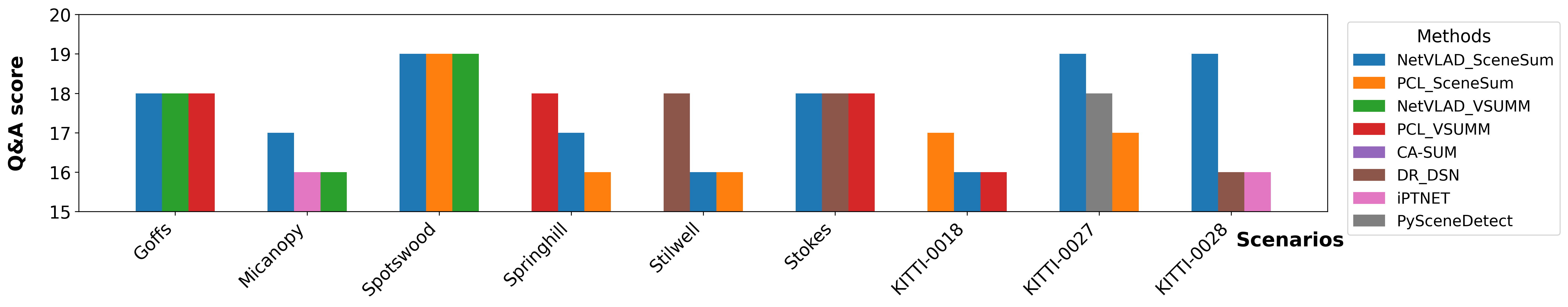}} \quad
    \vspace{-8mm}
    \caption{GPT-generated question answering performance on different scenes and methods (out of 20)}
    \label{fig:gpt-baseline}
\vspace{-2mm}
\end{figure*}

\textbf{Implementation details.}
The implementation for the KITTI dataset is similar to the one for the Habitat-Sim dataset, with the exception that we modify the distance threshold $r \in [0, 100]$ meters, to accommodate the scalability of the KITTI dataset. Furthermore, we carefully selected three scenes from the KITTI dataset for our training and testing procedures. Referring to our setup in the Habitat-Sim dataset, we also conducted the same cross-validation setup for the KITTI dataset. In this section, we only focus on self-supervised SceneSum.

\textbf{Baseline comparison.}
SceneSum performs the best which is especially remarkable considering KITTI's wide coverage. As in the last two rows of Tab.~\ref{tab:kitti}, SceneSum surpasses all baselines which validates the effectiveness of employing a clustering framework for scene summarization. Moreover, Fig.~\ref{fig:kitti_visualization} illustrates that akin to Habitat-Sim, the keyframes summarized by SceneSum are spatially sparse, distributed over a large area or range. SceneSum significantly surpasses all baselines, with AUC improvements of at least 34.3\%, 34.7\%, and 66.7\% in three scenes.





\vspace{-2mm}
\section{Downstream Tasks}\label{sec:downstream}
\vspace{-1mm}
\subsection{Q\&A}\label{sec:real-world indoor}
\vspace{-2mm}

To evaluate scene summarization methods on downstream tasks, we used GPT-4 to assess each method's support for question answering ($Q\&A$). This
 simulates applications requiring human-like scene understanding.

 The experiment has two main steps: First, GPT-4 generates questions based on a superset of all summarized images from each scene. Then, another GPT-4 prompt attempts to answer these questions using another set of images summarized by each method, without access to the initial question-generating images. Each method's performance is measured by the number of correct answers out of 20, reflecting the models' ability to capture the most informative and representative aspects of the scenes.

Fig.~\ref{fig:gpt-baseline} shows that SceneSum effectively generates diverse and detailed frame summaries, crucial for tasks like question answering ($Q\&A$). It operates within resource constraints while ensuring each frame captures unique and significant scene details, making it suitable for applications in surveillance, real estate, and robotics.

\vspace{-1mm}
\subsection{Sim2Real Transfer}\label{sec:sim2real}
\begin{figure}[t]
\vspace{-2mm}
    \centering
{\includegraphics[width=0.50\textwidth]{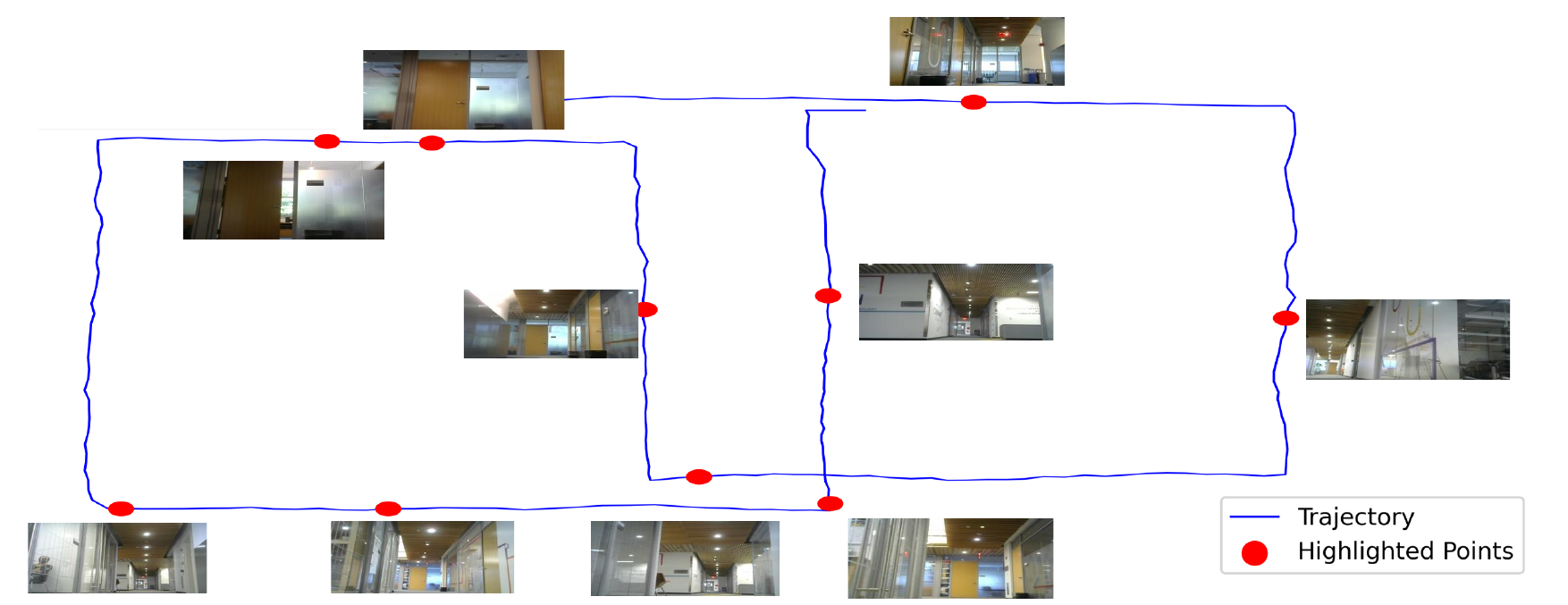}} \quad
    \vspace{-5mm}
    \caption{Selected
keyframes from real-time indoor dataset}

\label{fig:realtime_indoor_dataset}

\vspace{-6mm}

\end{figure}

In this section, we assess the generalization capabilities
of SceneSum, pretrained on a simulated dataset, using a real-world indoor
dataset. The real-world indoor dataset, collected by a robot with an RGB camera
and a trajectory obtained through surveying methods, consists of 300 RGB images
with a resolution of 640x480 pixels. The images are fed into the pretrained
MixVPR model to form 10 clusters, which are then processed by the pretrained
SceneSum model. The SceneSum model outputs 10 keyframes, one from each cluster.
Fig.~\ref{fig:realtime_indoor_dataset} highlights the keyframes selected from
the trajectory by the SceneSum model.

%% file: sections/5-conclusion.tex
\vspace{-2mm}
\section{Conclusion}\label{sec:conclusion}
\vspace{-1mm}

 In this study, we introduce a new task: scene summarization. We develop a two-stage supervised/self-supervised approach that combines autoencoders, contrastive learning, and ground truth supervision, enhancing keyframe selection by optimizing spatial sparsity and visual diversity. Additionally, we show that using Visual Place Recognition (VPR) for cluster acquisition significantly outperforms contrastive-based clustering methods.




%% file: sections/6-supplementary.tex
\clearpage
\section*{Supplementary}
\renewcommand{\thesection}{\Alph{section}}
\renewcommand{\thefigure}{\Roman{figure}}
\renewcommand{\thetable}{\Roman{table}}

\setcounter{section}{0}
\setcounter{figure}{0}
\setcounter{table}{0}

This supplement contains more ablation studies, time complexities, and visualizations that could not fit in the study. In particular, we include (1) extra examples on divergence vs distance threshold, (2) time complexity for different methods, (3) discussions on SLAM methods, and (4) more visualizations of selected keyframes.

\section{Extra examples on divergence vs distance threshold}
We present additional examples comparing divergence against distance threshold, comprising 4 examples on HabitatSim and 3 examples on KITTI. Overall, this follows the previous experiments conducted in Tab.~\ref{tab:time_habitat} and Tab.~\ref{tab:time_kitti}, where NetVLAD+SceneSum demonstrated the best results, followed by PCL+SceneSum.

\section{Time complexity for different methods}

We present the time taken by both the baseline methods and our approach in Tab.~\ref{tab:time_habitat} and Tab.~\ref{tab:time_kitti}. For non-learning-based methods, self-supervised, and supervised methods, we only consider the inference time for the test set. In contrast, for autolabeling, the training time is included in the total time calculation, as training is a crucial part of the optimization process.

\begin{table}[h]
\caption{Inference Time (seconds) for Habitat-Sim Dataset. (s) represents the methods supervised by ground truth. (a) represents the methods with autolabelling.}
\centering
\resizebox{1\linewidth}{!}{\begin{tabular}{|l|c|c|c|c|c|c|}
\hline
Scene & Goffs & Micanopy & Spotswood & Springhill & Stilwell & Stokes \\
\hline
DR-DSN & 63.376 & 27.983 & 30.089 & 37.623 & 38.099 & 22.257 \\
\hline
CA-SUM & 139.969 & 51.596 & 51.098 & 64.564 & 67.103 & 46.769 \\
\hline
PySceneDetect & 17.563 & 6.764 & 7.516 & 9.484 & 10.809 & 6.074 \\
\hline
VSUMM NetVLAD & 23.475 & 8.893  & 9.211 & 11.409 & 13.39 & 6.496 \\
\hline
VSUMM PCL & 20.03 & 9.702 & 9.85 & 13.002 & 12.156 & 7.713 \\
\hline
NetVLAD+SceneSum & 183.305 & 206.311 & 89.960 & 293.312 & 97.829 & 64.503 \\
\hline
PCL+SceneSum & 166.037 & 207.957 &  75.142 & 292.857 & 95.879 & 56.893 \\
\hline
NetVLAD+SceneSum(S) & 194.999 & 166.694 & 128.187 & 383.043 & 95.815 & 58.244 \\
\hline
PCL+SceneSum(S) & 185.233 & 172.785 &  147.617 & 394.343 & 84.859 & 60.510 \\
\hline
NetVLAD+SceneSum(A)& 148.894 & 149.157 & 81.360 & 304.567 & 78.026 & 54.432 \\
\hline
PCL+SceneSum(A)& 151.797 & 138.924 &  74.393 & 309.289 & 75.936 & 57.571 \\
\hline
\end{tabular}}
\label{tab:time_habitat}
\vspace{-5mm}
\end{table}

\vspace{-2mm}
\begin{table}[h]
\caption{Inference Time (seconds) for KITTI Dataset. Other abbreviations follow by Tab.~\ref{tab:time_habitat}}
\centering
\resizebox{0.9\linewidth}{!}{\begin{tabular}{|l|c|c|c|}
\hline
Scene & KITTI(0018) & KITTI(0027) & KITTI(0028)  \\
\hline
DR-DSN & 16.775 & 16.017 & 18.354 \\
\hline
CA-SUM & 52.493 & 34.428 & 43.639 \\
\hline
PySceneDetect & 1.92 & 4.015 & 4.48 \\
\hline
VSUMM NetVLAD & 1.78 & 3.098 & 3.461 \\
\hline
VSUMM PCL & 1.462 & 2.432 & 2.898 \\
\hline
NetVLAD+SceneSum & 26.395 & 43.996 & 60.562  \\
\hline
PCL+SceneSum & 22.730 & 37.828 & 56.860  \\
\hline
NetVLAD+Supervised & 19.084 & 38.316 & 56.531  \\
\hline
PCL+Supervised & 18.183 & 34.909 & 52.305 \\
\hline
NetVLAD+Autolabelling & 17.350 & 34.002 & 50.562  \\
\hline
PCL+Autolabelling & 17.442 & 33.426 & 46.654  \\
\hline
\end{tabular}}
\label{tab:time_kitti}
\vspace{-5mm}

\end{table}


\section{Discussions on SLAM methods}
\begin{figure}[h]
  \centering
  \setlength{\abovecaptionskip}{4pt}
    \centering
     \includegraphics[width=0.9\linewidth]{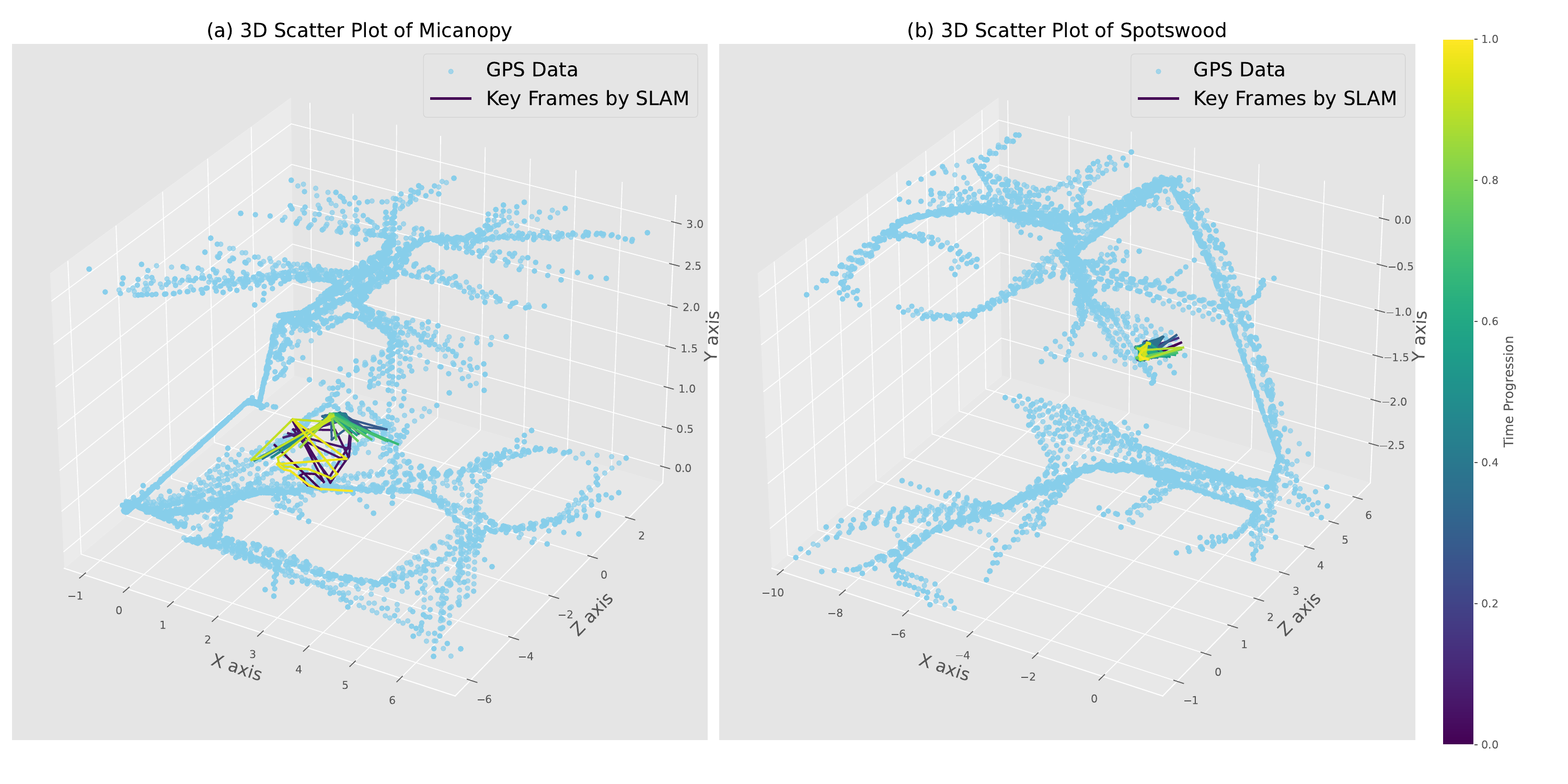}
    \caption{\textbf{OpenVSLAM fails to create a complete map in Scene Micanopy and Spotswood}}
    \label{fig:SLAM}
\vspace{-2mm}
\end{figure}

Some may argue that SLAM+KNN could easily solve this problem. However, attempts to map entire scenes using SLAM methods like OpenVSLAM were unsuccessful in Habitat environments at a sampling rate of 1 frame per second. In our study, OpenVSLAM faced a major challenge with tracking loss, preventing the generation of a complete map, as shown in Fig.~\ref{fig:SLAM}.
\vspace{-1mm}

We employ OpenVSLAM as one of our SLAM (Simultaneous Localization and Mapping) methods, but it frequently experiences tracking loss during the map generation phase, leading to disorganized and unstructured maps. This recurring issue highlights a fundamental limitation of our current SLAM approach, suggesting the need for further investigation and refinement to achieve reliable and well-structured map generation.

\begin{figure}[h]
  \centering
  \setlength{\abovecaptionskip}{3pt}
    \includegraphics[width=0.92\linewidth]{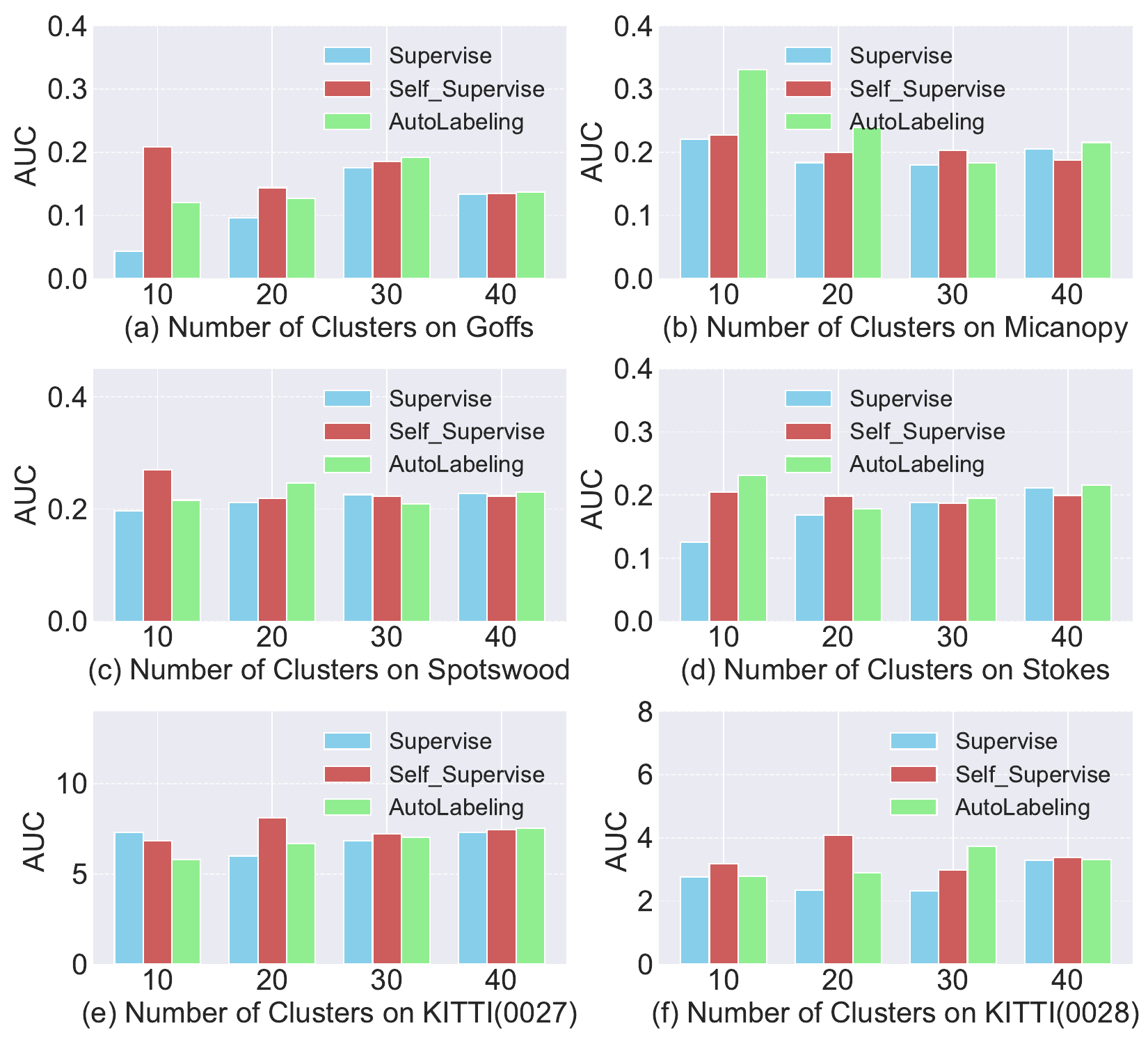}
    \caption{Performance Analysis of Supervised Method, Self-Supervised Method and AutoLabeling under Different Number of Clusters in (a) Goffs (b) Micanopy (c) Spotswood (d) Springhill (e) KITTI(0027) (f) KITTI(0028)}
    \label{fig:SupervisedvsSelf-supervised}
\end{figure}
\vspace{-2mm}

\begin{figure}[h]
    \centering
    \begin{subfigure}{0.23\textwidth}
        \centering
        \includegraphics[width=\textwidth]{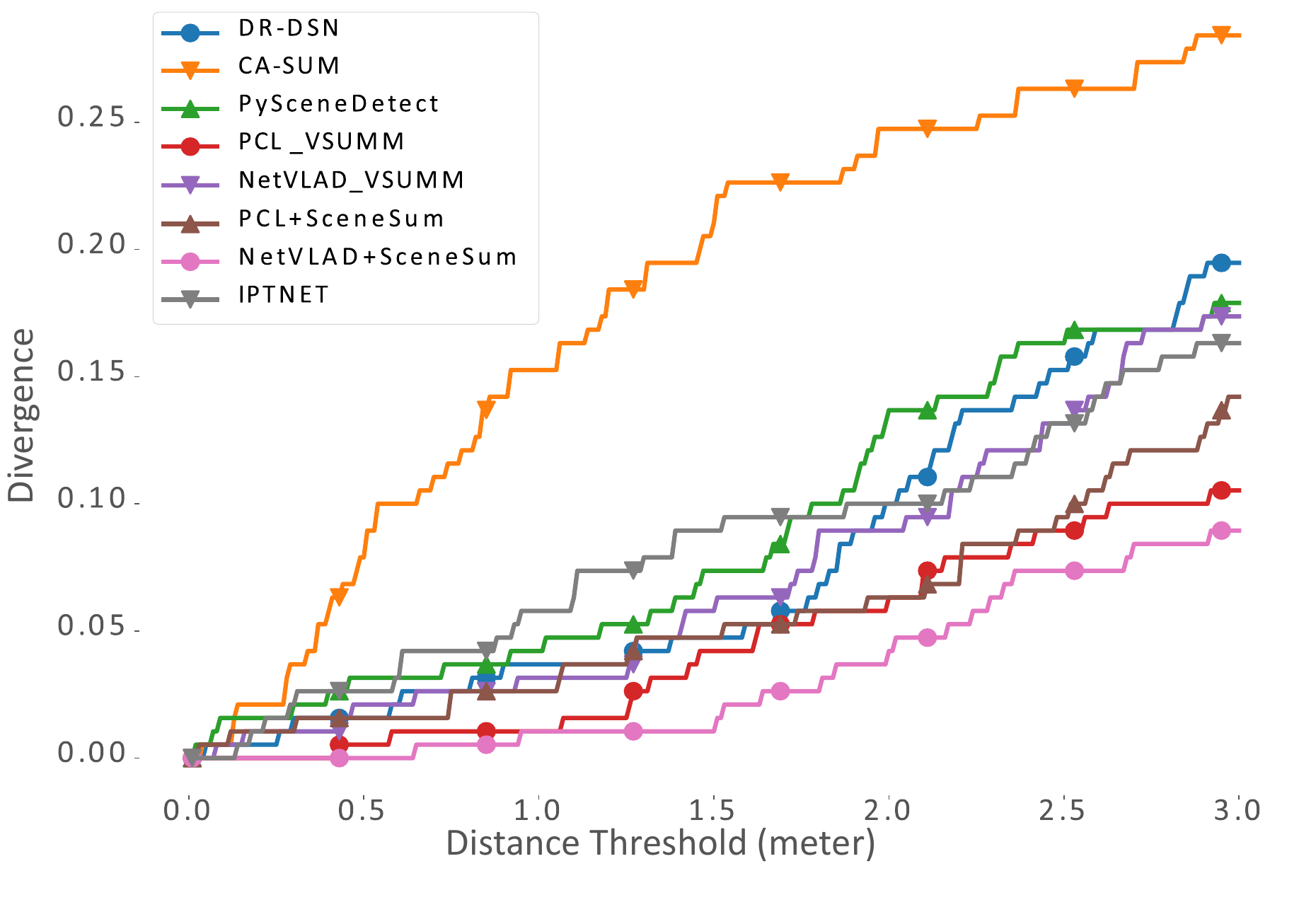}
        \caption{Habitat-Sim (Stilwell) at 20 Summarized Frames}
        \label{fig:div_habitat}
    \end{subfigure}
    \begin{subfigure}{0.23\textwidth}
        \centering
        \includegraphics[width=\textwidth]{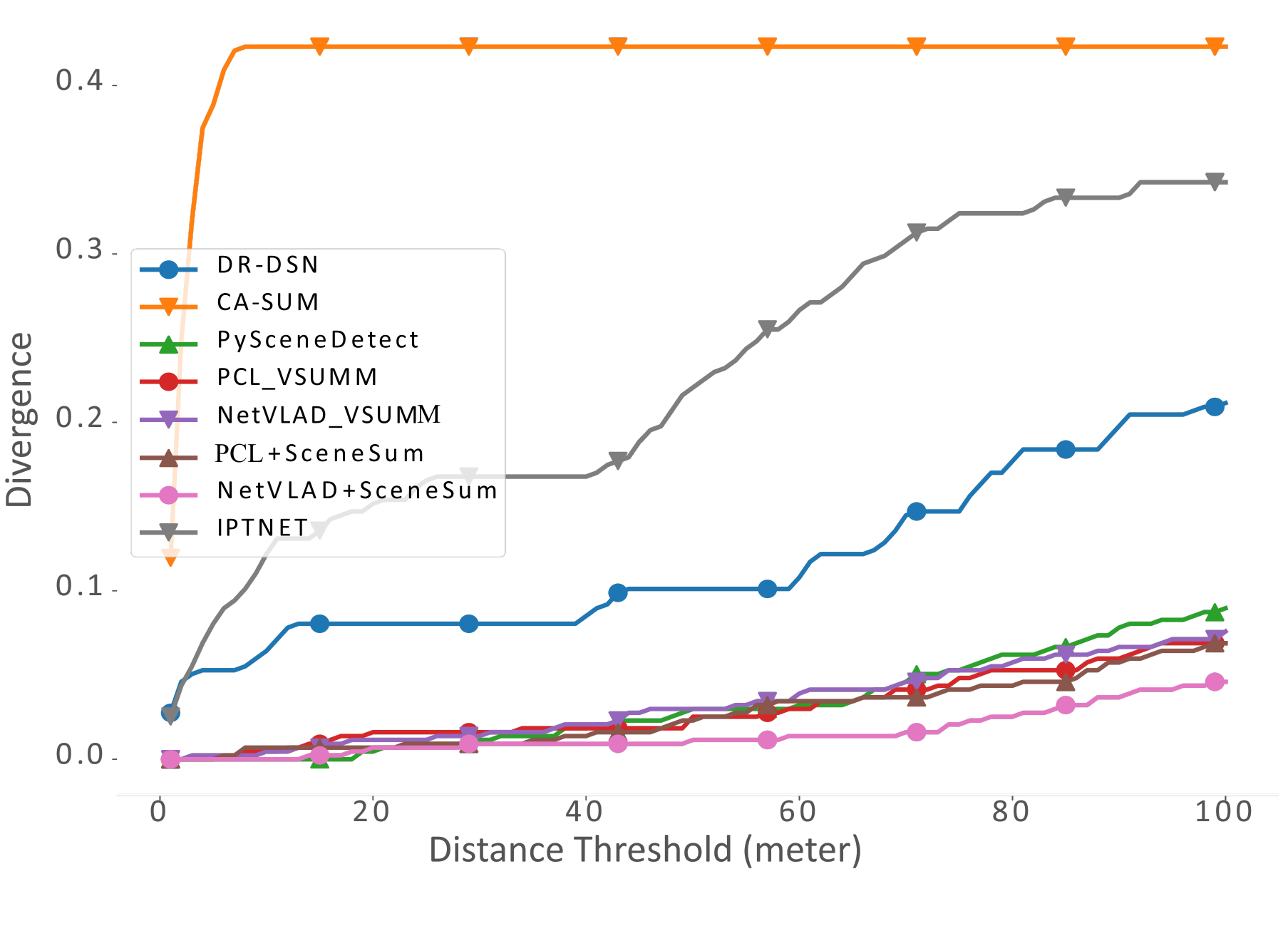}
        \caption{KITTI (0028) at 30 Summarized Frames.}
        \label{fig:div_KITTI}
    \end{subfigure}
    \vspace{-3mm}
    \caption{Comparison of Divergence vs Distance Threshold for Habitat-Sim and KITTI dataset}
    \vspace{-5mm}
    \label{fig:div_comparison}
    
\end{figure}

\subsection{Ablation Study}\label{sec:ablation_study}

\textbf{Supervised vs Self-supervised.} This section aims to determine if introducing supervision loss results in significant performance changes in the model. Fig.~\ref{fig:SupervisedvsSelf-supervised} presents the average performance of supervised and self-supervised methods on the Habitat-Sim and KITTI environments. It compares these two models across six scenarios at cluster sizes of 10, 20, 30, and 40, using the AUC metric. Generally, the supervised model slightly outperforms the self-supervised model. Notably, with 10 clusters, the supervised approach shows a substantial improvement over the self-supervised one. For other cluster sizes, incorporating ground truth data during training offers only a slight performance boost. This highlights the robustness and effectiveness of self-supervised SceneSum across various contexts, proving it to be a adaptable tool, especially in situations where ground truth data is unavailable.


\begin{figure*}[h]
    \centering
    \begin{subfigure}{0.25\textwidth}
        \centering
        \includegraphics[width=\linewidth]{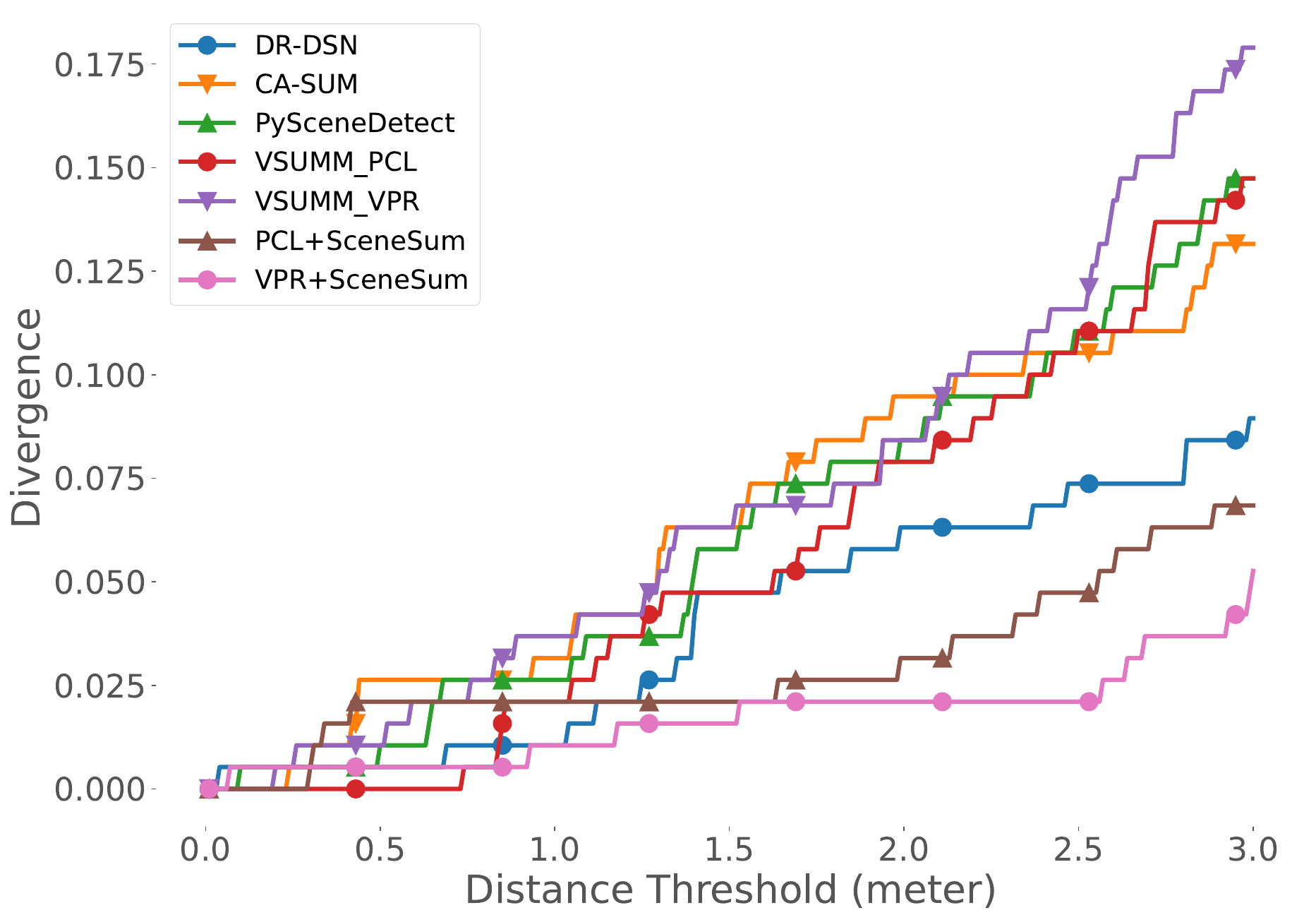}
        \caption{Habitat-Sim (Goffs)}
        \label{fig:div_goff}
    \end{subfigure}
    \hfill
    \begin{subfigure}{0.25\textwidth}
        \centering
        \includegraphics[width=\linewidth]{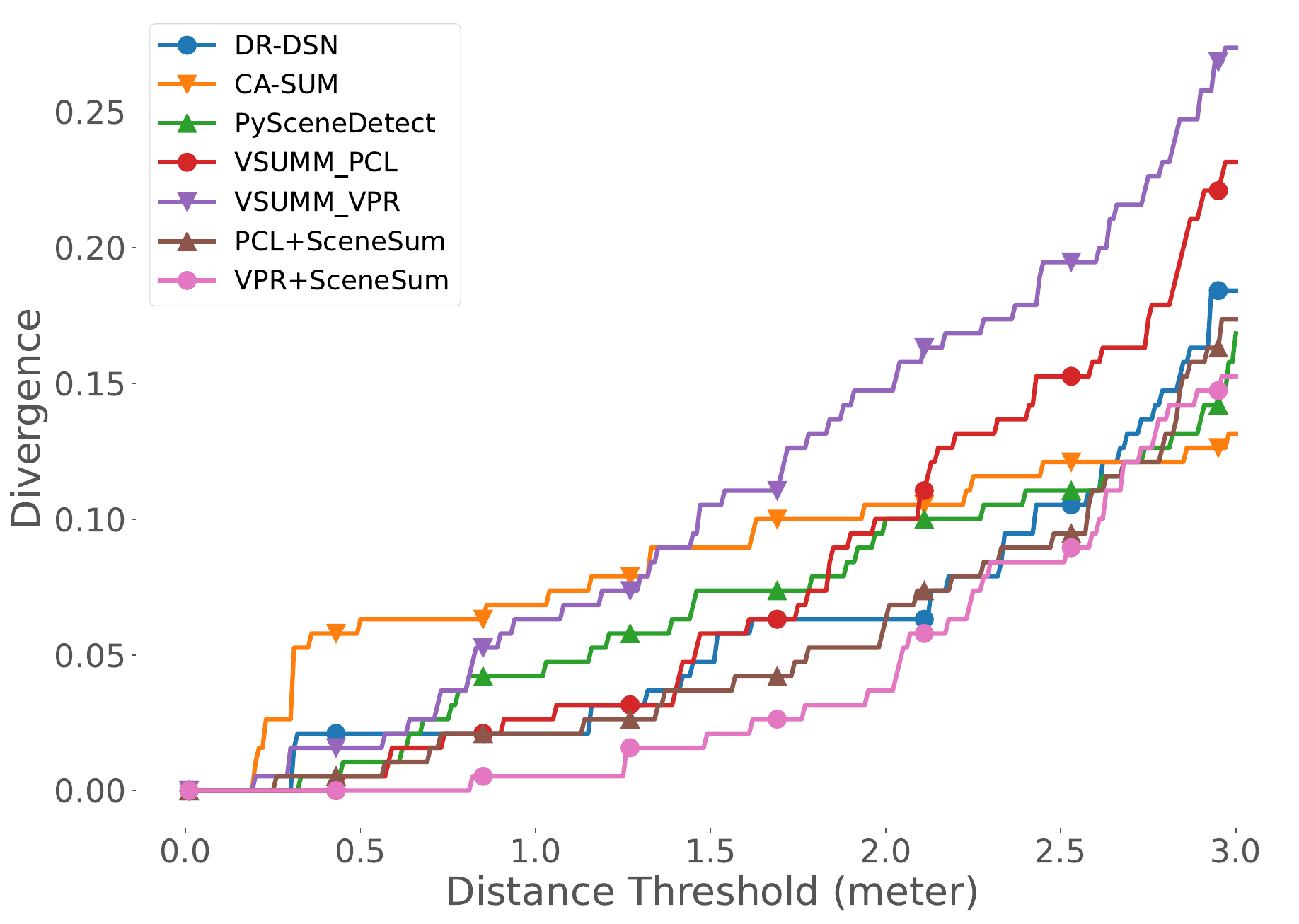}
        \caption{Habitat-Sim (Micanopy)}
        \label{fig:div_micanopy}
    \end{subfigure}
    \hfill
    \begin{subfigure}{0.25\textwidth}
        \centering
        \includegraphics[width=\linewidth]{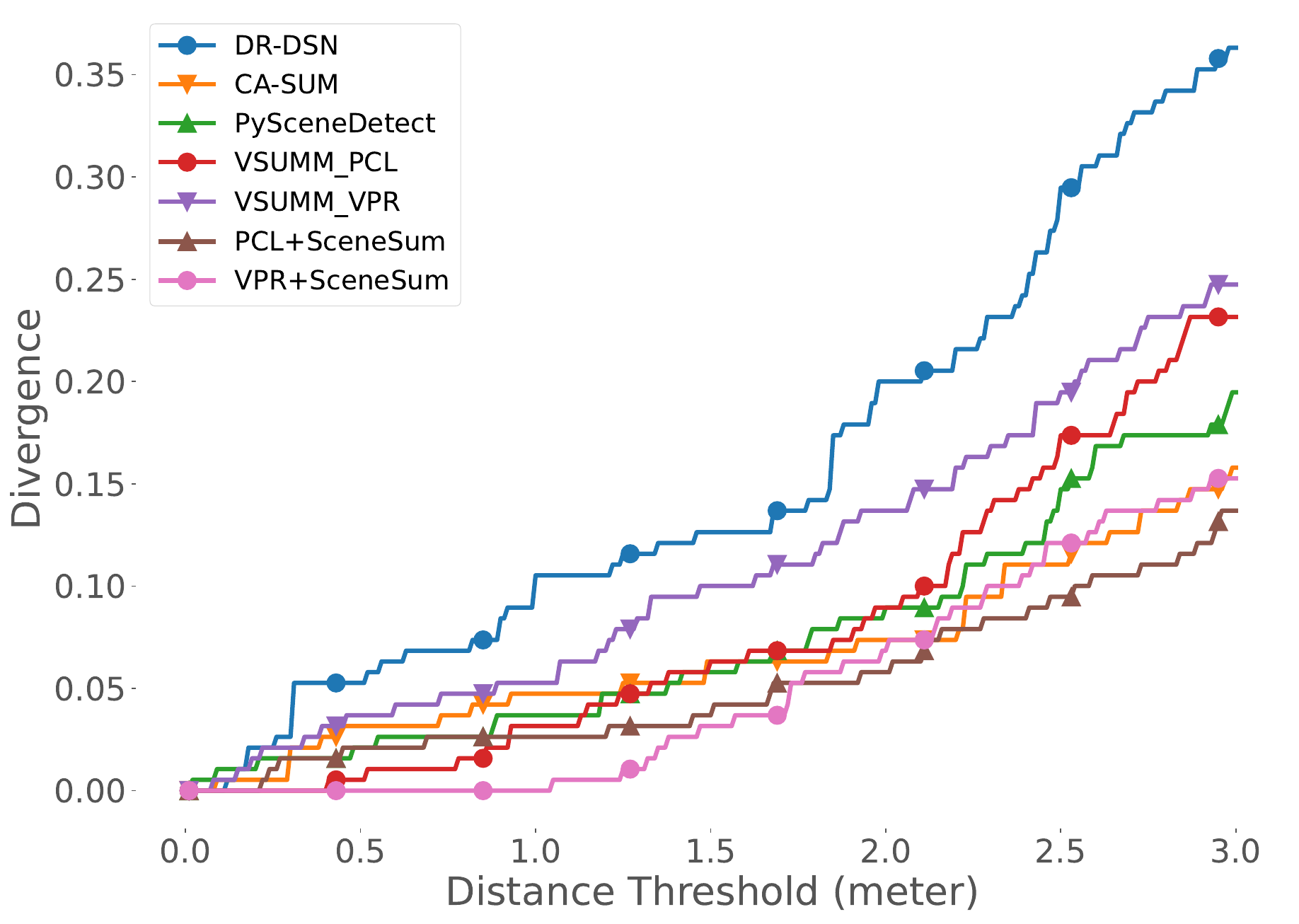}
        \caption{Habitat-Sim (Spotswood)}
        \label{fig:div_Spotswood}
    \end{subfigure}
    
    \begin{subfigure}{0.25\textwidth}
        \centering
        \includegraphics[width=\linewidth]{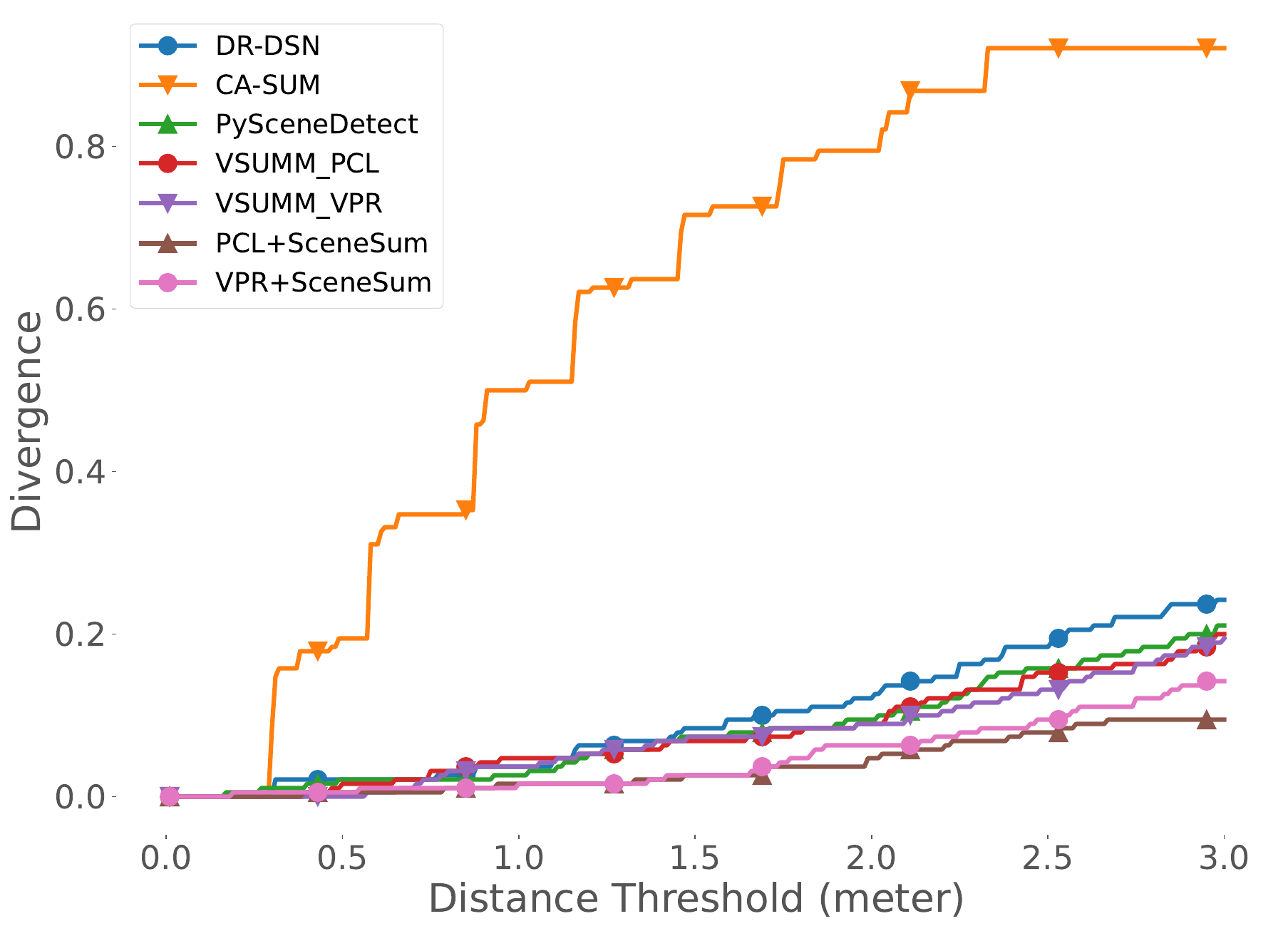}
        \caption{Habitat-Sim (Stokes)}
        \label{fig:div_Stokes}
    \end{subfigure}
    \hfill
    \begin{subfigure}{0.25\textwidth}
        \centering
        \includegraphics[width=\linewidth]{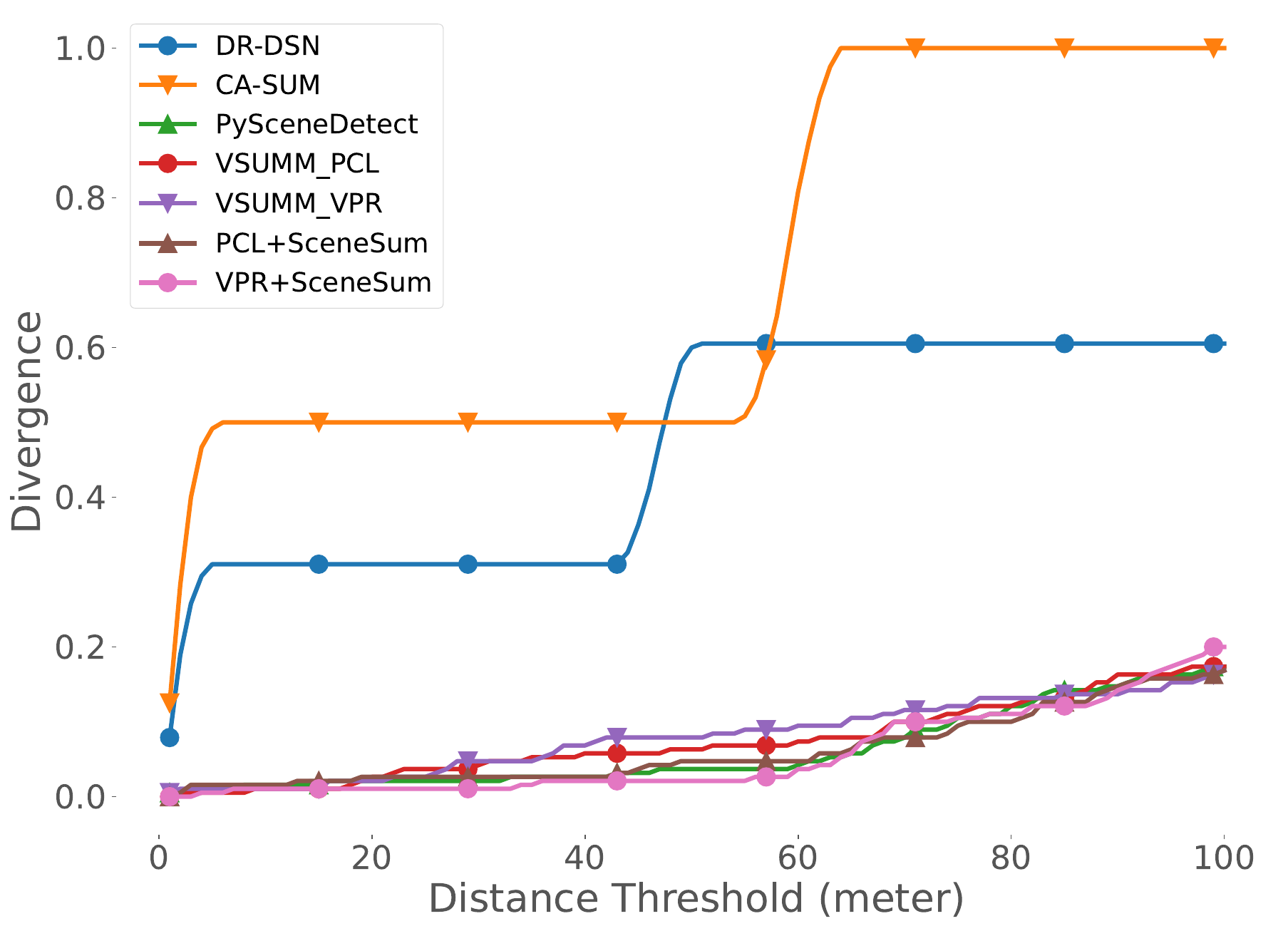}
        \caption{KITTI (0018)}
        \label{fig:div_0018}
    \end{subfigure}
    \caption{Divergence vs Distance Threshold at 20 Summarized Frames}
    \label{fig:div_2x3}
\vspace{-3mm}
\end{figure*}


\textbf{More VPR vs Contrastive-based clustering comparison.} As discussed in Sec.~\ref{sec:ab_cluster},
Tab.~\ref{tab:habitat} and Tab.~\ref{tab:kitti} again show that SceneSum, when combined with VPR-based clustering, significantly outperforms contrastive-based clustering in most scenes. This advantage is due to VPR's ability to capture distinctive information about locations and places, focusing on scene context and spatial relationships. In contrast, the contrastive-based clustering method captures only visual information, making VPR-based clustering more suitable for scene summarization.

\textbf{Divergence comparisons.} Fig.~\ref{fig:div_habitat} and~\ref{fig:div_KITTI} highlight a significant trend in the performance of SceneSum approaches for scene summarization: as the distance threshold increases, the divergence also rises across all methods. NetVLAD+SceneSum consistently outperforms all baseline methods at every distance threshold, showing stable performance regardless of the distance evaluated. Additional results for more scenes are presented in Fig.~\ref{fig:div_2x3}


\begin{table}[ht]
\centering
\setlength{\abovecaptionskip}{3pt}
\vspace{-3mm}
\caption{Best AUC results for different clustering on Habitat-sim}
\label{tab:vpr}
\resizebox{1\linewidth}{!}{
\begin{tabular}{|c|c|c|c|c|c|c|c|c|}
\hline
\diagbox[width=6.4em]{Clustering}{Scene} & \textbf{Goffs} & \textbf{Micanopy} & \textbf{Spotswood} & \textbf{Springhill} & \textbf{Stilwell} & \textbf{Stokes} & \textbf{AVG.} & \textbf{SD.} \\ \hline
NetVLAD & \textcolor{black}{0.050} & \textcolor{black}{0.117} & \textcolor{black}{0.146} & \textcolor{black}{0.092} & \textcolor{black}{0.111} & \textcolor{black}{0.135} & \textcolor{black}{0.109} & \textcolor{black}{0.031} \\ 
\hline
MixVPR & \textcolor{black}{0.084} & \textcolor{black}{0.179} & \textcolor{black}{0.194} & \textcolor{black}{0.094} & \textcolor{black}{0.084} & \textcolor{black}{0.140} & \textcolor{black}{0.129} & \textcolor{black}{0.045} \\
\hline
Patch & \textcolor{black}{0.053} & \textcolor{black}{0.098} & \textcolor{black}{0.121} & \textcolor{black}{0.102} & \textcolor{black}{0.073} & \textcolor{black}{0.119} & \textcolor{black}{0.094} & \textcolor{black}{0.024} \\ 
\hline
\end{tabular}
}
\vspace{-2mm}
\end{table}

    
 Sec.~\ref{sec:habitat_experiment},~\ref{sec:kitti_experiment}, and~\ref{sec:ablation_study} reveal a key finding: self-supervised SceneSum exhibits strong summarization performance when applied zero-shot to a new scene. In this section, we explore the model's auto-labeling capability. If the time required for scene summarization is not a concern, SceneSum can function as a network summarizer, summarizing the scene video while simultaneously training on it.

Fig.~\ref{fig:SupervisedvsSelf-supervised} shows that the auto-labeling results closely match those of the self-supervised approach, confirming the effectiveness of self-supervised inference. Even when comparing self-supervised SceneSum to a version that overfits the test dataset by training on it, the outcomes remain similar.

\section{More visualizations on selected keyframes}
\begin{figure*}[!h]
  \centering
  \setlength{\abovecaptionskip}{4pt}
  \begin{minipage}{1\textwidth} 
    \centering
    \includegraphics[width=1.0\linewidth]{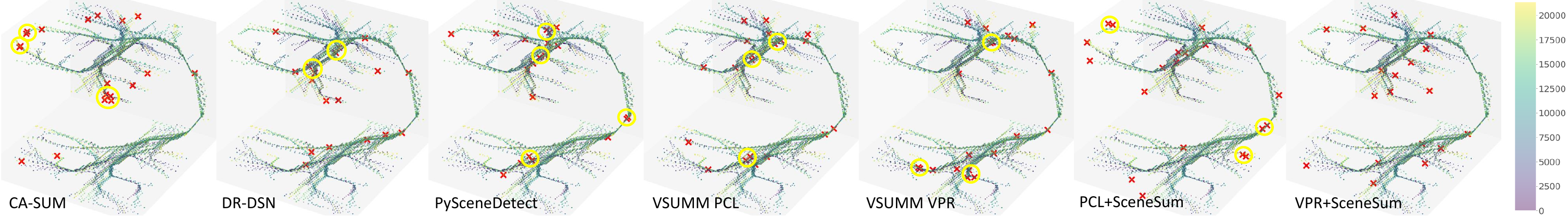}
    \caption{\textbf{Selected keyframes in Habitat-Sim Dataset.} We summarize 20 keyframes of 7 baselines on scene \textit{Goffs}. All frames are color-coded by temporal order. Summarized keyframes are marked with red crosses. Groups of frames that are geographically close to each other are circled in yellow.}
    \label{fig:goffs_visualization}
  \end{minipage}\hfill
  \vspace{3mm}

  \centering
  \setlength{\abovecaptionskip}{4pt}
  \begin{minipage}{1\textwidth} 
    \centering
    \includegraphics[width=1.0\linewidth]{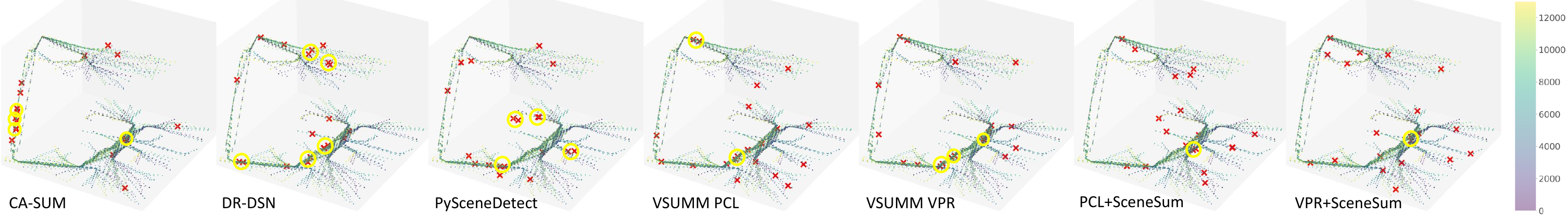}
    \caption{\textbf{Selected keyframes in Habitat-Sim Dataset.} We summarize $20$ keyframes of $7$ baselines on scene \textit{Stilwell}. The baselines and annotations follow Fig.~\ref{fig:goffs_visualization}
    }
    \label{fig:stilwell_visualization}
    \vspace{3mm}
  \end{minipage}\hfill

  \centering
  \setlength{\abovecaptionskip}{4pt}
  \begin{minipage}{1\textwidth} 
    \centering
    \includegraphics[width=1.0\linewidth]{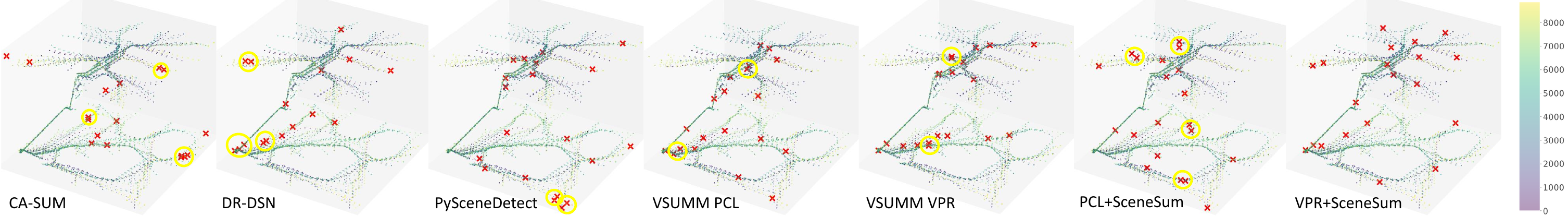}
    \caption{\textbf{Selected keyframes in Habitat-Sim Dataset.} We summarize $20$ keyframes of $7$ baselines on scene \textit{Micanopy}. The baselines and annotations follow Fig.~\ref{fig:goffs_visualization}}
    \label{fig:micanopy_visualization}
  \end{minipage}\hfill

  \centering
  \setlength{\abovecaptionskip}{4pt}
  \begin{minipage}{1\textwidth} 
    \centering
    \includegraphics[width=1.0\linewidth]{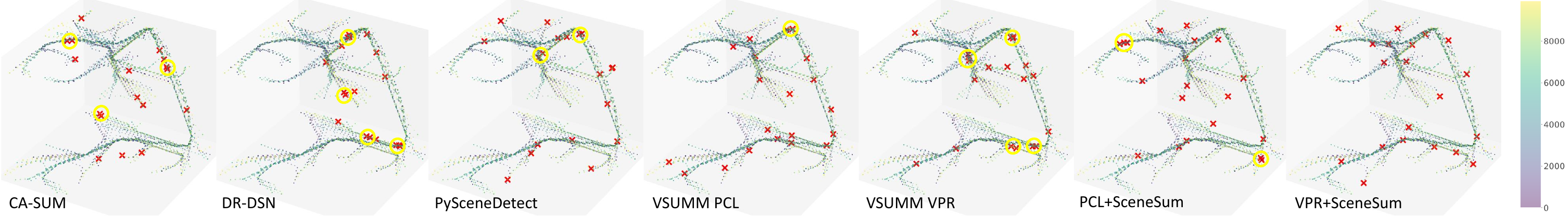}
    \caption{\textbf{Selected keyframes in Habitat-Sim Dataset.} We summarize $20$ keyframes of $7$ baselines on scene \textit{Spotswood}. The baselines and annotations follow Fig.~\ref{fig:goffs_visualization}}
    \label{fig:spotswood_visualization}
  \end{minipage}\hfill
  \vspace{3mm}

  \centering
  \setlength{\abovecaptionskip}{4pt}
  \begin{minipage}{1\textwidth} 
    \centering
    \includegraphics[width=1.0\linewidth]{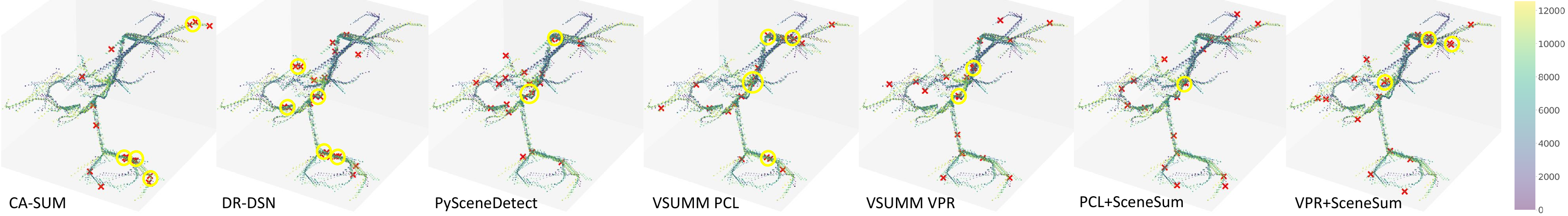}
    \caption{\textbf{Selected keyframes in Habitat-Sim Dataset.} We summarize $20$ keyframes of $7$ baselines on scene \textit{Springhill}. The baselines and annotations follow Fig.~\ref{fig:goffs_visualization}}
    \label{fig:springhill_visualization}
  \end{minipage}\hfill
  \vspace{3mm}

  \centering
  \setlength{\abovecaptionskip}{4pt}
  \begin{minipage}{1\textwidth} 
    \centering
    \includegraphics[width=1.0\linewidth]{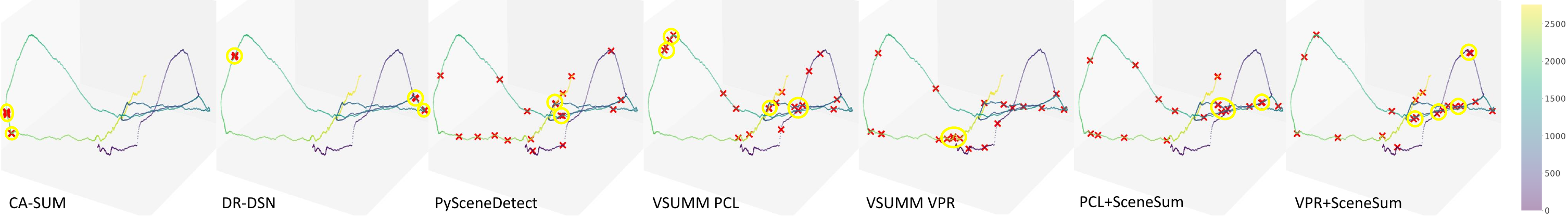}
    \caption{\textbf{Selected keyframes in KITTI Dataset.} We summarize $20$ keyframes of $7$ baselines on scene \textit{0018}. The baselines and annotations follow Fig.~\ref{fig:goffs_visualization}}
    \label{fig:0018_visualization}
  \end{minipage}\hfill
  \vspace{3mm}

  \centering
  \setlength{\abovecaptionskip}{4pt}
  \begin{minipage}{1\textwidth} 
    \centering
    \includegraphics[width=1.0\linewidth]{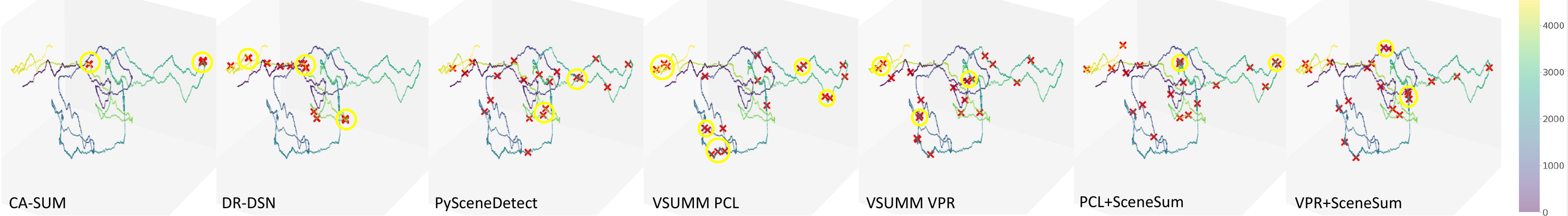}
    \caption{\textbf{Selected keyframes in KITTI Dataset.} We summarize $20$ keyframes of $7$ baselines on scene \textit{0027}. The baselines and annotations follow Fig.~\ref{fig:goffs_visualization}}
    \label{fig:0027_visualization}
  \end{minipage}\hfill
  \vspace{3mm}
\end{figure*}

We have presented more visualizations on selected keyframes, including $5$ Habitat scenes and $2$ KITTI scenes, as shown in Figure.~\ref{fig:goffs_visualization}, Figure.~\ref{fig:stilwell_visualization}, Figure.~\ref{fig:micanopy_visualization}, Figure.~\ref{fig:spotswood_visualization}, Figure.~\ref{fig:springhill_visualization}, Figure.~\ref{fig:0018_visualization}, and Figure.~\ref{fig:0027_visualization}.

